\documentclass[runningheads,a4paper]{llncs}
\usepackage{amssymb}
\usepackage{graphicx}
\usepackage{marvosym}
\usepackage{booktabs} 
\usepackage{amsmath}
\usepackage{url}
\usepackage{hyperref}
\usepackage{makecell}
\usepackage{tabularx}
\usepackage{booktabs}
\usepackage{geometry}
\usepackage{etoolbox}
\usepackage{xcolor}
\pretocmd{\bibliography}{\begingroup\scriptsize}{}{}
\apptocmd{\bibliography}{\endgroup}{}{}

\apptocmd{\thebibliography}{\scriptsize}{}{}
\newcommand{\keywords}[1]{\par\addvspace\baselineskip
\noindent\keywordname\enspace\ignorespaces#1}
\begin{document}

\mainmatter  

\title{Adversarial Examples in Environment Perception for Automated Driving}

\titlerunning{Lecture Notes in Computer Science: Authors' Instructions}

%
%
\author{Jun Yan%
\thanks{This review is a section of the ``Automated Driving Vehicle Technologies" book.}%
\and Huilin Yin\Letter}
%

\institute{School of Electronic and Information Engineering, Tongji University, No. 4800, Caoan Gonglu Road, Shanghai, China\\}

%
%

\toctitle{SOTIF Guarantee Technologies for Autonomous Vehicles}
\tocauthor{Authors' Instructions}
\maketitle

\begin{abstract}
The renaissance of deep learning has led to the massive development of automated driving. However, deep neural networks are vulnerable to adversarial examples. The perturbations of adversarial examples are imperceptible to human eyes but can lead to the false predictions of neural networks. It poses a huge risk to artificial intelligence (AI) applications for automated driving. This survey systematically reviews the development of adversarial robustness research over the past decade, including the attack and defense methods and their applications in automated driving. The growth of automated driving pushes forward the realization of trustworthy AI applications. This review lists significant references in the research history of adversarial examples.
\end{abstract}

\section{Introduction}
\indent 
\par Deep learning has been hugely successful over the past decade, powered by Graph Processing Units (GPUs), big data, and human intelligence. This success has spurred an artificial intelligence (AI) renaissance, enabling amazing applications like chat assistants, embodied robotics, and autonomous driving. In the past decade, AI technologies have matured enough for autonomous driving to be productized. The Society of Automotive Engineers (SAE) categorizes driving levels from no automation (Level 0) to full automation (Level 5), aiming for fully autonomous driving \cite{SAETerm}. With AI advancements, some countries have introduced decrees for testing fully autonomous systems.
 \begin{figure}[!h]
        \centering
\includegraphics[width=0.5\hsize]{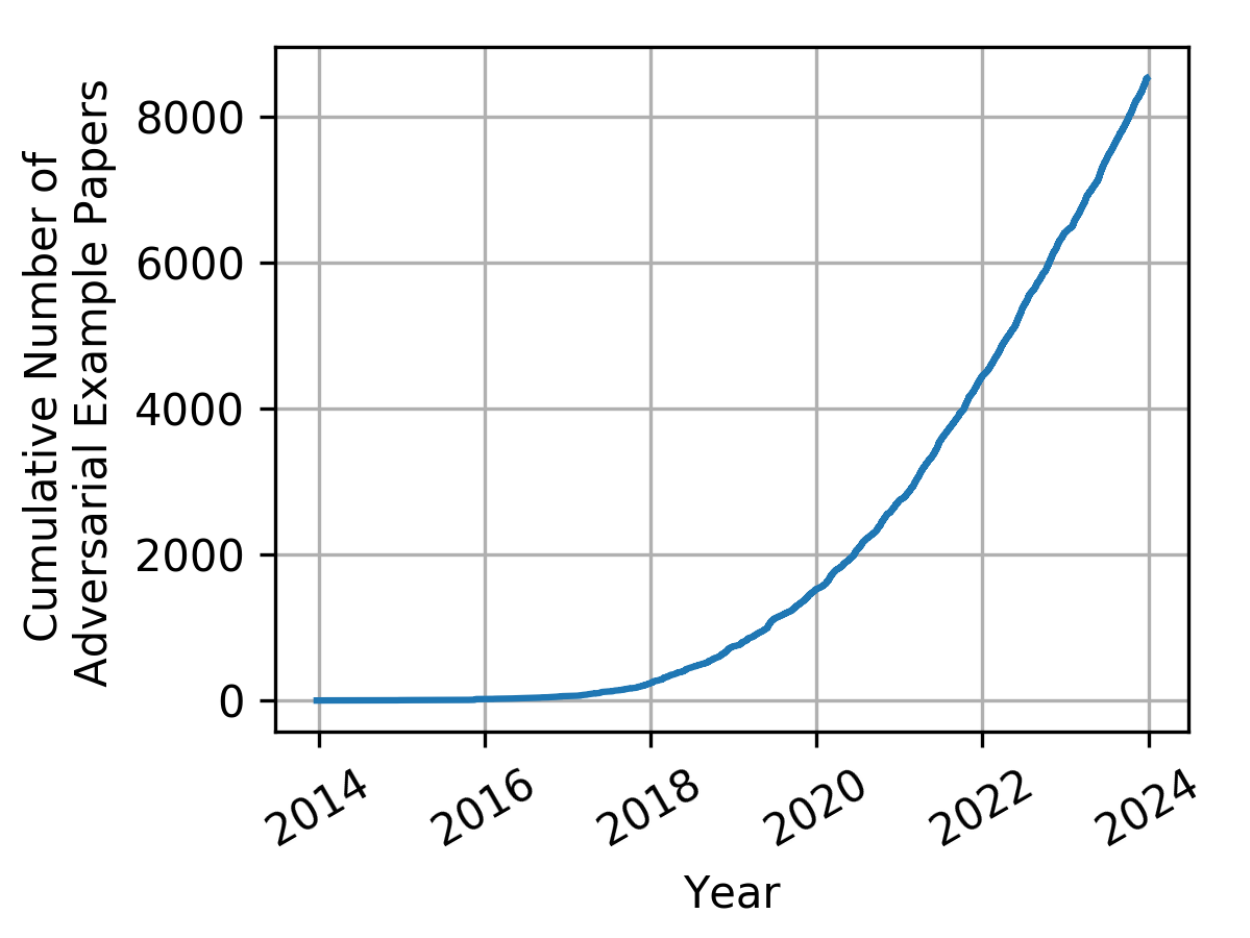}
\caption{The growth of papers related to adversarial robustness~\cite{carlini2019blog} @Nicholas Carlini's Blog.}
\label{fig:adv_example_growth}
\end{figure}
\par However, deep neural networks are vulnerable to adversarial attacks \cite{szegedy2014intriguing,goodfellow2015explaining,carlini2017towards,madry2018towards} where slight raw data perturbations fool networks into wrong predictions. Many pieces of research have delved into the exploration to promote AI security, including attacks \cite{szegedy2014intriguing,goodfellow2015explaining,carlini2017towards}, defenses \cite{madry2018towards,zhang2019theoretically}, systematic evaluations \cite{dong2020benchmarking,dong2023benchmarking,tang2021robustart}, and interpretations \cite{yin2019fourier,fawzi2016robustness}. Fig. \ref{fig:adv_example_growth} shows huge research interest in adversarial robustness, although valuable scientific problems may be saturated. This saturation validates the impact and significant of the adversarial robustness research.
\par
\begin{figure}[!t]
        \centering
\includegraphics[width=1.0\hsize]{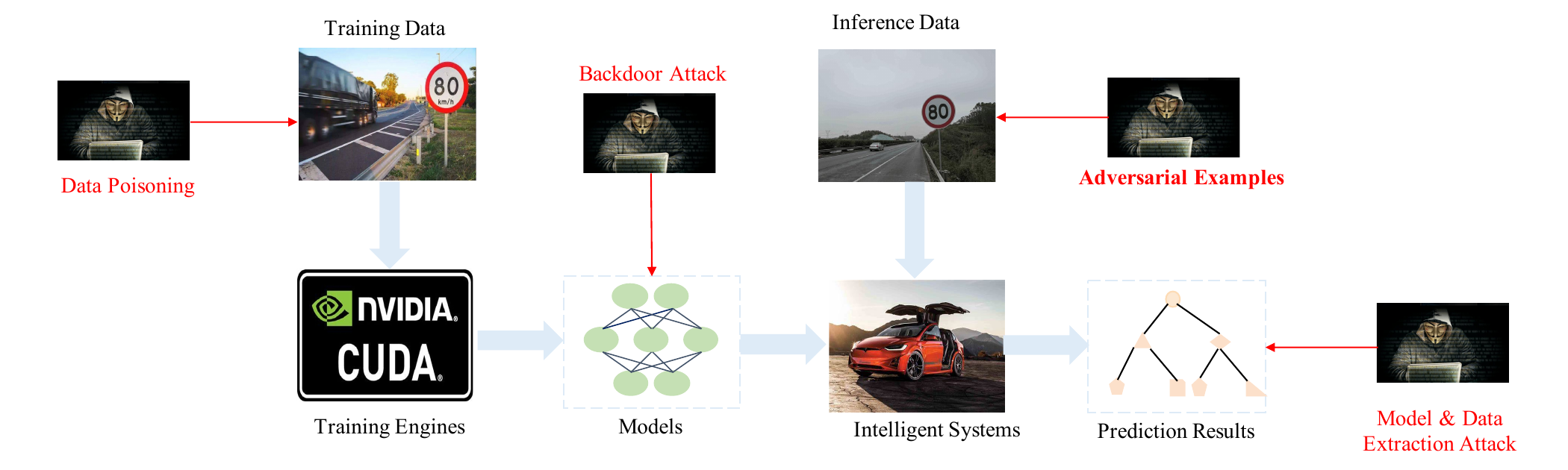}
\caption{Deep learning systems and the encountered attacks. Adversarial attacks happen in the model prediction process.}
\label{fig:macro_story_attack_draw}
\end{figure}

 The issue of adversarial examples relates to the security risks of cyber-physical systems. In these systems, adversarial examples can serve as malicious data in the Internet of Vehicles. For instance, an automated vehicle's decision-making system can be misled by maliciously modifying a traffic sign. These adversarial examples reveal AI systems' vulnerabilities, inducing wrong decisions and potential security risks. Understanding adversarial examples' characterization and creation is critical for designing secure cyber-physical systems and ensuring the information systems' robustness. Meanwhile, other security and privacy risks also harm AI technologies' trustworthiness. Fig. \ref{fig:macro_story_attack_draw} shows a macro story of AI technology risks. Adversarial examples primarily threaten during the inference phase of AI systems. Despite the importance of paying attention to other AI risks, this survey mainly reviews the research history of adversarial examples.
\par
\begin{figure}[!t]
        \centering
\includegraphics[width=0.85\hsize]{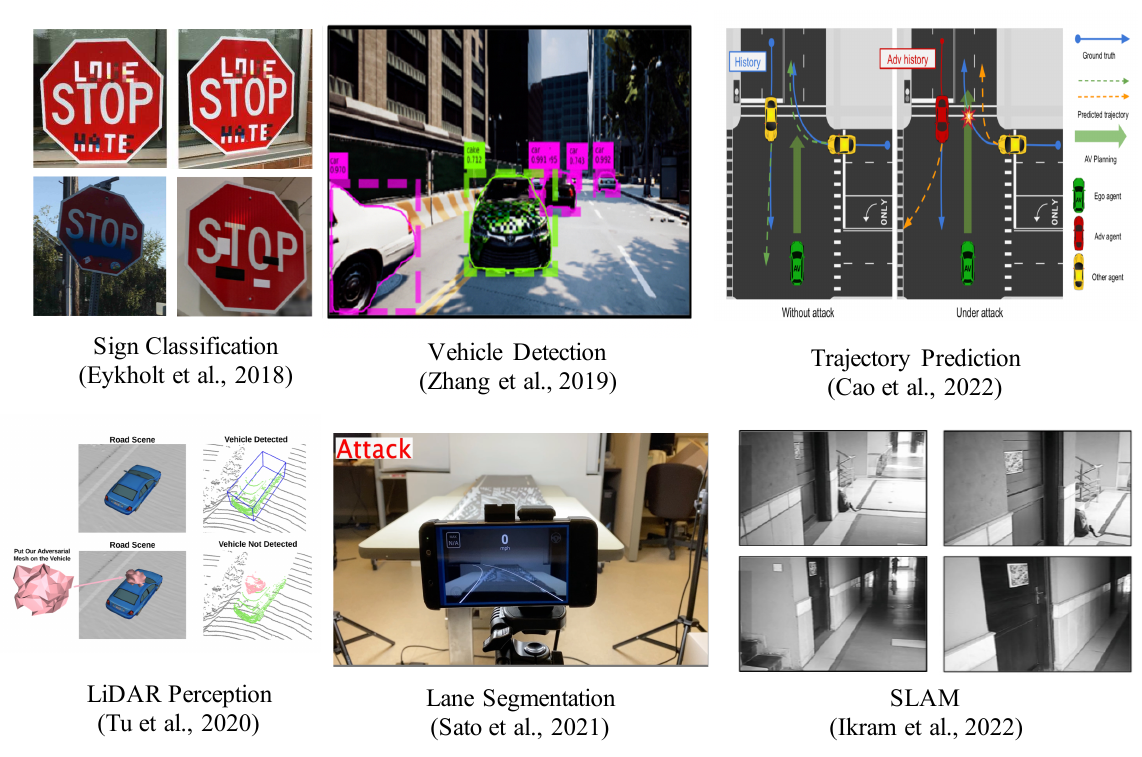}
\caption{The threat of adversarial examples in the practical tasks of automated driving.}
\label{fig:physical_attack_scenario}
\end{figure}

 Giving attention to adversarial robustness research is critical to advancing the development of intelligent vehicles. The existence of adversarial examples poses a huge threat to the tasks of automated driving, including traffic sign recognition~\cite{eykholt2018robust}, vehicle detection~\cite{zhang2019camou}, trajectory prediction~\cite{zhang2022adversarial}, LiDAR perception~\cite{cao2019adversarial}, lane segmentation~\cite{sato2021dirty}, and SLAM (Simultaneous Localization and Mapping)~\cite{ikram2022perceptual}. Fig. \ref{fig:physical_attack_scenario} demonstrates the adversarial vulnerability risk towards the automated driving, which indicates the necessity to advance the research on adversarial robustness.
\par This review introduces adversarial examples' theories, methods, and automated driving applications. In particular, it focuses on the adversarial examples related to the environment perception systems for automated driving. Section 2 interprets fundamental concepts. Section 3 describes representative adversarial attack methods. Section 4 recalls adversarial defense methods. Section 5 introduces adversarial examples in automated driving applications. Section 6 highlights adversarial examples' relationship to the Safety of the Intended Functionality (SOTIF). Section 7 provides a future outlook for adversarial robustness research. Finally, Section 8 concludes the review.

\section{Theory Preliminary}
\begin{table}[!t]

	\caption{Notations \& Explanations}
	\label{tab: notation}
	\begin{tabular*}{\linewidth}{@{}cl@{}}
		\toprule[1.5pt]
		Symbol & \multicolumn{1}{c}{Description} \\ \midrule[1pt]
		$x_{0}$ & A clean sample \\
		$\tilde{x_{i}}$ & An adversarial example \\
            $\mathbf{x}$ & The vectors of clean data \\
            $\mathbf{y}$ & The vectors of labels \\
            $\mathbf{\tilde{x}}$ & The vectors of adversarial examples \\
		$\epsilon$ & The perturbation budget \\
		$\mathcal{B}(x_0,\epsilon)$ & The neighborhood ball of $x_0$ with radius of $\epsilon$ \\
		$x^{\prime}$ & A sample in $\mathcal{B}(x_0,\epsilon)$ \\
            $\mathbf{x^{\prime}}$ & The vector sample in $\mathcal{B}(x_0,\epsilon)$ \\
		$y_0$ & The groundtruth label \\
		$\mathbf{I}$ & The unit matrix \\
            $\nabla_{z} J(\cdot)$ & The gradient of a scalar function $J$ with respect to $\mathbf{z}$ \\
		$f_{\mathbf{\Theta}}$ & The neural network function with the parameter space $\mathbf{\Theta}$ \\
		$\delta$ & The added perturbation \\
		$\|\boldsymbol{x}\|_{p}(p \geq 1)$ & The vector $p$-norm of $\boldsymbol{x}=\left[x_{1} \mid, \ldots, x_{d}\right]$, defined as $\|x\|_{p}=\left(\sum_{i=1}^{d}\left|x_{i}\right|^{p}\right)^{1 / p}$\\
            $\|x\|_{\infty}$ & infinity norm of $\boldsymbol{x}=\left[x_{1} \mid, \ldots, x_{d}\right]$, defined as $\|x\|_{\infty}=\max _{i \in[d]}\left|x_{i}\right|$\\
		$\mathcal{L}(\boldsymbol{\cdot})$ & The adversarial loss function \\
            $K$ & The number of classes in a classification task \\
		$L$ & The Lipschitz constant \\
		$\odot$ & Hadamard product \\ \bottomrule[1.5pt]
	\end{tabular*}
\end{table}
\indent 
\par This section reviews the essential concepts and theory preliminary in the adversarial examples research. Table \ref{tab: notation} gives essential notations which would be utilized to illustrate the studies. The following subsections review the mechanisms of gradient-based adversarial attacks, adversarial training, and randomized smoothing.
\subsection{Gradient-based Adversarial Attacks}
\indent 
\par Given the original examples $\mathbf{x_{0}}$, a successful adversary aims to find the relative adversarial examples $\mathbf{\tilde{x}}=\mathbf{{x}_{0}}+\mathbf{\Delta}$ that can deceive the visual system with the small perturbation $\mathbf{\Delta}$. It is a constrained optimization problem that the adversarial examples locate in the norm sphere of original examples defined in Eq. (\ref{eq:norm_ball_constraint}):
\begin{equation}\label{eq:norm_ball_constraint}
\underset{\mathbf{\tilde{x}}}{\operatorname{argmax}} f(\mathbf{\tilde{x}}) \text { s.t. } \mathbf{\tilde{x}} \in \mathcal{B}\left(\mathbf{x}_{0}\right),
\end{equation}
where $f(\mathbf{\tilde{x}})$ denotes the deceit on the classifier function $f(.)$ and $\mathcal{B}\left(\mathbf{x}_{0}\right)$ is a small
region (norm sphere) with the adversarial perturbations. The norm sphere $\mathcal{B}\left(\mathbf{x}_{0}\right)$ is also a constrained set that the $\ell_{p}$ norm can measure the distance defined in Eq. (\ref{eq:norm_distance}):
\begin{equation}\label{eq:norm_distance}
\mathcal{B}\left(\mathbf{x}_{0}\right)=\left\{\mathbf{\tilde{x}}:\left\|\mathbf{\tilde{x}}-\mathbf{x}_{0}\right\|_{p} \leq \epsilon\right\}.
\end{equation}
\par An original groundtruth label is $y_{0}$. There exist two types of attack categories. If the adversarial example $\tilde{x}_{i}$ belongs to a specific class $y_{t}$, this is a targeted attack defined in Eq. (\ref{eq:targeted_attack}):
\begin{equation}\label{eq:targeted_attack}
\underset{{\tilde{x}_{i}}}{\operatorname{argmax}} f({\tilde{x}_{i}}) = y_{t}.
\end{equation}
\textcolor{black}{This means that the adversary induces the model to misclassify the data as the specific wrong label.} Otherwise, it is an untargeted attack defined in Eq. (\ref{eq:untargeted_attack}):
\begin{equation}\label{eq:untargeted_attack}
\underset{{\tilde{x}_{i}}}{\operatorname{argmax}} f({\tilde{x}_{i}}) \neq y_{0}.
\end{equation}
\textcolor{black}{In such a scenario, the adversary induces the model to misclassify the data as the unspecific wrong label.}
\par The gradient-based attack is a $\ell_{\infty}$-norm steepest descent attack. The adversary utilizes a linear approximation of the objective function to search for the perturbations. Assumed that $g(.)$ is an attack procedure, $\mathbf{x}$ and $\mathbf{x^{\prime}}$ is a batch of clean data and the searched examples in the neighborhood of $\mathbf{x}$. Eq. (\ref{eq:taylor_approximation}) defines the linear approximation of first-order Taylor expansion. 
\begin{equation}\label{eq:taylor_approximation}
g(\mathbf{x^{\prime}}) \approx g\left(\mathbf{x}\right)+\nabla g\left(\mathbf{x}\right)^{T}\left(\mathbf{x^{\prime}}-\mathbf{x}\right).
\end{equation}
A closed-form solution defined in Eq. (\ref{eq:fgsm})of this constrained optimization problem can be derived:
\begin{equation}\label{eq:fgsm}
\mathbf{x}^{\prime}=\mathbf{x}+\epsilon \operatorname{sign}\left(\nabla g\left(\mathbf{x}\right)\right).
\end{equation}
The symbol $\operatorname{sign}(\cdot) \in\{+1,-1\}$ denotes element-wise sign values. Eq. (\ref{eq:fgsm}) is a mathematical form of the fast gradient sign method (FGSM), which is a milestone work of AI security. \textcolor{black}{The FGSM attack pipeline can realize both untargeted and targeted attacks.} If the adversary runs the FGSM method for multiple iterations $T$, Eq. (\ref{eq:pgd}) can be deduced:
\begin{equation}\label{eq:pgd}
\boldsymbol{x}_{t+1}=\Pi\left(\boldsymbol{x}_{t}+\alpha \operatorname{sign}\left(\nabla g\left(\boldsymbol{x}_{t}\right)\right)\right) \quad \forall t=0, \ldots, T-1,
\end{equation}
where $\alpha$ is a step size for gradient-based attack and $\mathbf{x}_{t}$ is the perturbed data in the time step $t$. The multiple-step attack would terminate if the sampled data of adversarial examples $\mathbf{\tilde{x}}$ is found. Eq. (\ref{eq:pgd}) defines the mathematical equation of iterative FGSM (I-FGSM) attack~\cite{kurakin2018adversarial} and projected gradient descent (PGD) attack~\cite{madry2018towards}. Compared to the I-FGSM attack, the PGD attack projects the perturbed input back onto the set of allowable inputs to ensure the modified image still has pixel values in the valid range. Currently, the PGD attack is a de facto effective gradient-based attack in the practical application.
\subsection{Adversarial Defense}
 \begin{figure}[!t]
        \centering
\includegraphics[scale=0.35]{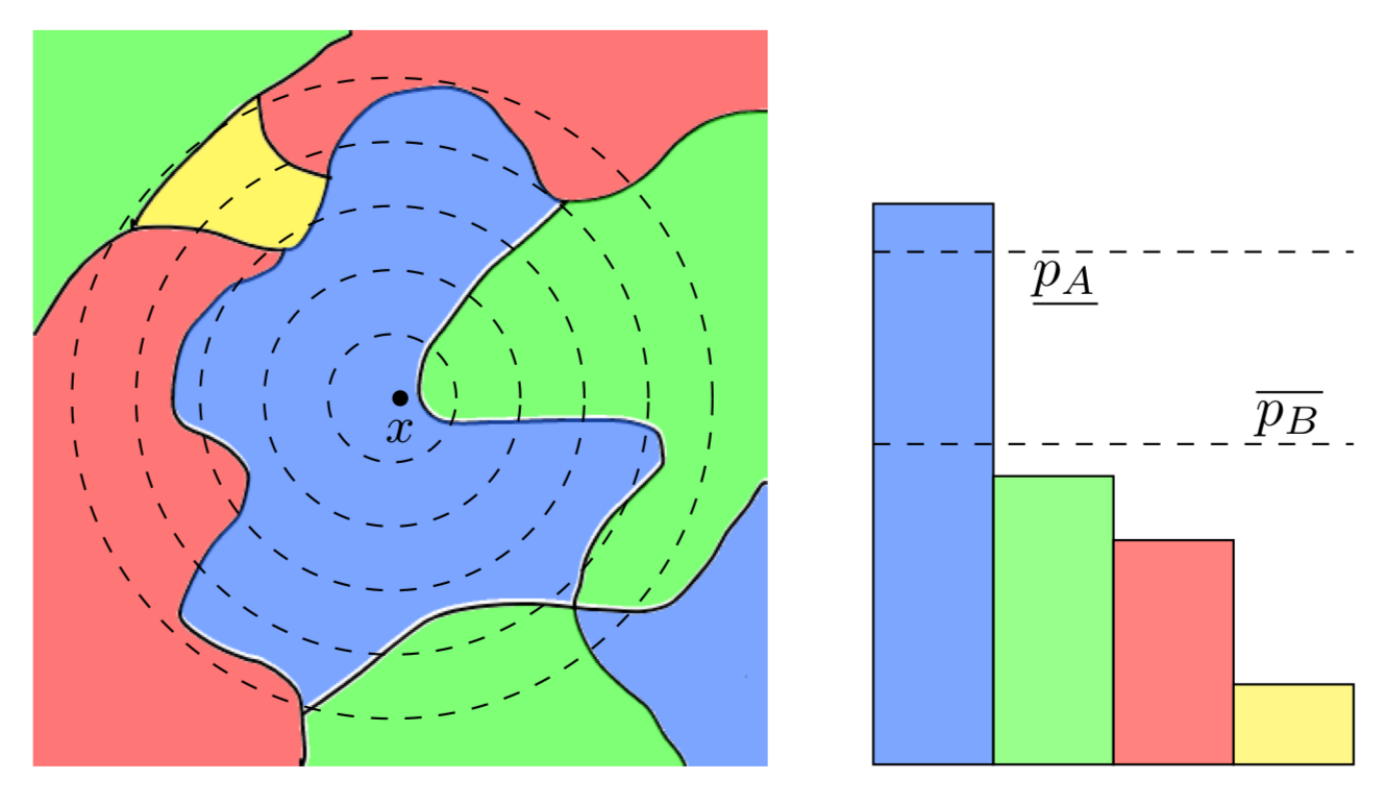}
\caption{The evaluation paradigm of smoothed classifier~\cite{cohen2019certified}. \textbf{Left}: the decision regions of the base classifier $f$ are marked in different colors. The dotted lines represent the level sets of the distribution $\mathcal{N}\left(x, \sigma^{2} I\right)$. \textbf{Right}: the distribution $\mathcal{N}\left(x, \sigma^{2} I\right)$. Here, $\underline{p_{A}}$ is a lower bound on the probability of the top class, and $\overline{p_{B}}$ is an upper bound on the probability of each other class. The classifier function is in blue.}
\label{fig:randomized_smoothing}
\end{figure}
\indent
\par The Section of Theory Preliminary would mainly focus on two effective defense methods under the adversarial attacks: adversarial training~\cite{madry2018towards} and certified randomized smoothing~\cite{cohen2019certified}. Other methods will be introduced in the next paragraphs.
\par The adversarial training procedure aims to minimize the expected empirical risk while maximizing the adversarial perturbations. Eq. (\ref{eq:adv_training}) defines the form of adversarial training:
\begin{equation}\label{eq:adv_training}
\arg \min _{\theta} E_{(\mathbf{x}, \mathbf{y}) \in \mathcal{D}} \max _{\mathbf{x^{\prime}} \in\left\|\mathbf{x^{\prime}}-\mathbf{x}\right\|_{\infty} \leq \epsilon} L\left({f_{\mathbf{\theta}}}\left(\mathbf{x^{\prime}}\right), \gamma\right),
\end{equation}
where $E$ denotes expectation, $(\mathbf{x}, \mathbf{y}) \in \mathcal{D}$ denotes the data samples and their labels randomly drawn from the distribution $\mathcal{D}$, $L(.)$ is a supervised loss function, $f_{\mathbf{\theta}}$ is the fitting function of neural networks, and $\mathbf{x^{\prime}}$ is the symbol of the perturbed data of $\epsilon$-ball. Adversarial training is an effective shield against the adversarial attacks to handle the crisis of model leakages or transfer-based attacks.
\par Beside the empirical defense such as adversarial training, the defenders also desire to compute the ``certified radius"~\cite{cohen2019certified}, where it provides a robustness guarantee with a high probability that any perturbation within such radius will give a robust prediction. Fig. \ref{fig:randomized_smoothing} describes the mechanism of certified randomized smoothing. Assumed that $g(.)$ is a smoothed classifier with the base classifier $f(.)$ under the adversarial attacks, where the data is perturbed by isotropic Gaussian noise, there exists a formulation defined in Eq. (\ref{eq:randomized_smoothing_eq1}): 
\begin{equation}\label{eq:randomized_smoothing_eq1}
\begin{array}{l}\qquad g(x)=\underset{c \in \mathcal{Y}}{\arg \max } \mathbb{P}(f(x+\varepsilon)=c) \\ \text { where } \varepsilon \sim \mathcal{N}\left(0, \sigma^{2} I\right)\end{array}.
\end{equation}
\par Assume that when the base classifier $f(.)$ categorizes a sample drawn from the distribution $\mathcal{N}\left(x, \sigma^{2} I\right)$, it returns the most probable class $c_{A}$ with the probability $p_{A}$. Simultaneously, the second most probable class, referred to as the ``runner-up," is returned with a probability of $p_{B}$. The smoothed classifier is robust within the $\ell_{2}$ radius $R=\frac{\sigma}{2}\left(\Phi^{-1}\left(p_{A}\right)-\Phi^{-1}\left(p_{B}\right)\right)$, where $\Phi^{-1}$ is the inverse of the standard Gaussian CDF (Cumulative Distribution Function). This review will highlight two important theorems of certified randomized smoothing.
\par Theorem \ref{thm:randomized_smoothing_thm1} is a crucial result. The certified robustness can be built on the neural network models, if any, are satisfied by modern deep architectures. The certified radius $R$ tends to be large under the following conditions: (1) the noise level $\sigma$ is high; (2) the probability associated with the top class $c_{A}$ is high; and (3) the probabilities corresponding to all other classes are low.
\begin{theorem}\label{thm:randomized_smoothing_thm1}~\cite{cohen2019certified}
Let $f: \mathbb{R}^{d} \rightarrow \mathcal{Y}$ be be any deterministic or
random function, and let $\varepsilon \sim \mathcal{N}\left(0, \sigma^{2} I\right)$. Let g be defined as in Eq. (\ref{eq:randomized_smoothing_eq1}). Suppose $c_{A} \in \mathcal{Y} \text { and } \underline{p_{A}}, \overline{p_{B}} \in[0,1]$ satisfy:
\begin{equation}\label{eq:randomized_smoothing_eq2}
\mathbb{P}\left(f(x+\varepsilon)=c_{A}\right) \geq \underline{p_{A}} \geq \overline{p_{B}} \geq \max _{c \neq c_{A}} \mathbb{P}(f(x+\varepsilon)=c).
\end{equation}
Then $g(x+\delta)=c_{A}$ for all $\|\delta\|_{2}<R$, where
\begin{equation}\label{eq:randomized_smoothing_eq3}
R=\frac{\sigma}{2}\left(\Phi^{-1}\left(\underline{p_{A}}\right)-\Phi^{-1}\left(\overline{p_{B}}\right)\right).
\end{equation}
\end{theorem}
\par Theorem \ref{thm:randomized_smoothing_thm2} demonstrates that Gaussian smoothing inherently leads to $\ell_{2}$ robustness. Specifically, if the assumptions about the base classifier are limited solely to class probabilities as in Eq. (\ref{eq:randomized_smoothing_eq2}), then the range of perturbations against which a Gaussian-smoothed classifier can be provably defended aligns precisely with an $\ell_{2}$ ball.
\begin{theorem}\label{thm:randomized_smoothing_thm2}~\cite{cohen2019certified}
Assume $\underline{p_{A}}+\overline{p_{B}} \leq 1$. For any perturbation $\delta$ with $\|\delta\|_{2}>R$, there exists a base classifier $f(.)$ consistent
with the class probabilities defined in in Eq. (\ref{eq:randomized_smoothing_eq2}) for which $g(x+\delta) \neq c_{A}$.
\end{theorem}
\par Adversarial training and randomized smoothing are two promising directions of defense. The following sections will introduce these two methods and other significant defense methods.
\section{Adversarial Attacks}
 \begin{figure}[!t]
        \centering
\includegraphics[width=1.0\hsize]{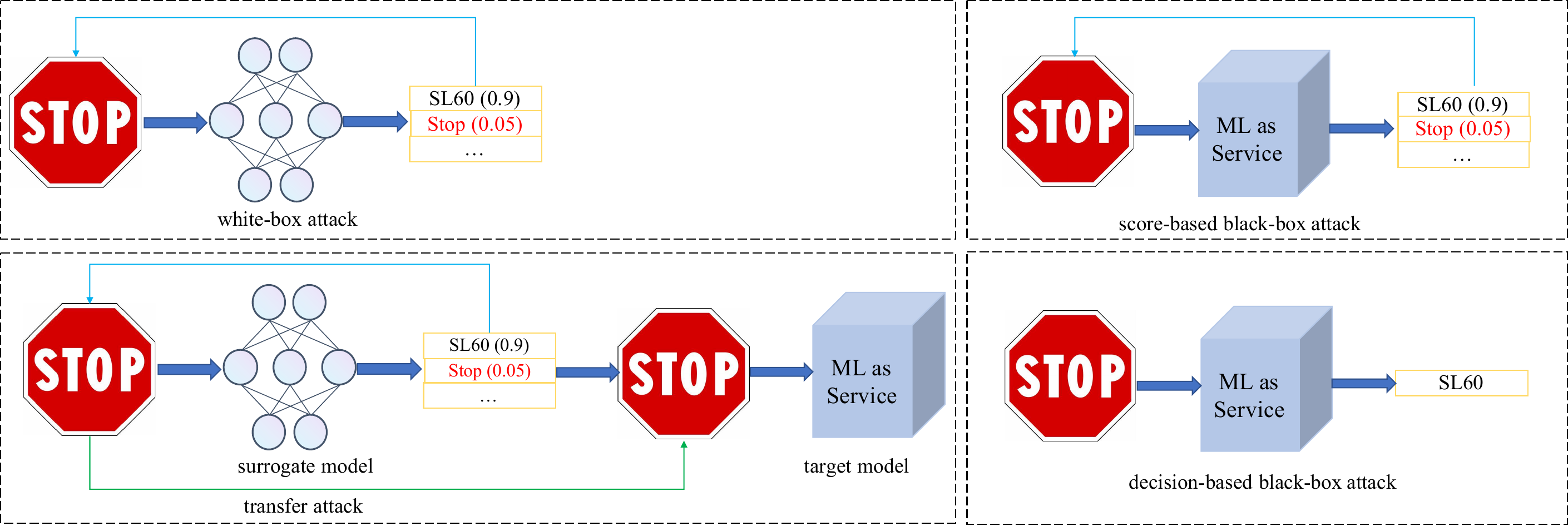}
\caption{Taxonomy and illustration of different categories of adversarial attacks.}
\label{fig:attack_category}
\end{figure}
\indent
\par Adversarial attacks represent a significant security threat to artificial intelligence (AI) systems, manifesting primarily during the model prediction phase, as depicted in Figure \ref{fig:macro_story_attack_draw}. This section delves into various adversarial attack methodologies, highlighting their diversity and impact.
\par According to the degree of knowledge and mastery of modeling, adversarial attacks can be categorized as the generation of white-box adversarial examples and black-box adversarial examples. Fig. \ref{fig:attack_category} illustrates the categories of different attacks. In the white-box adversarial attacks, the adversaries can obtain the model knowledge to launch the invasions. In black-box adversarial attacks, the adversaries would carry out a sneak attack on the Machine Learning as a Service (MLaaS) system that the model information is protective to the users. When an attacker obtains a white-box adversarial example with a high attack success rate (ASR) on a source model and migrates it to a target model for an attack, man calls it a black-box transfer attack. The MLaaS systems will provide an output under the black-box adversarial attacks. If the output is a confidence vector, the attack is a score-based adversarial attack. If the output is a specific category, the attack is a decision-based adversarial attack.
\par Table \ref{tab:digital_attack_methods} presents a compilation of seminal digital attack methods developed over the past decade, reflecting the significant scholarly contributions in this domain. The ensuing subsections will elaborate on these methods, providing a comprehensive understanding of their mechanisms and implications.
\begin{table}[!t]
\scriptsize
\centering
\caption{Classical adversarial attack methods. The main text gives the full names of abbreviations.}
\label{tab:digital_attack_methods}
\begin{tabularx}{\textwidth}{@{} l @{\hspace{5pt}} *{5}{X} @{}}
\toprule
Method & Distance & Physical attack & Knowledge & Iterative & Targeted \\
\midrule
L-BFGS~\cite{szegedy2014intriguing} & $\ell_{2}$ & No & White & Yes & Yes \\
FGSM~\cite{goodfellow2015explaining} & $\ell_{\infty}$ & No & White & No & No \\
BIM~\cite{kurakin2016adversarial} & $\ell_{\infty}$ & Yes & White & Yes & No \\
PGD~\cite{madry2018towards} & $\ell_{\infty}$ & No & White & Yes & No \\
MIM~\cite{dong2018boosting} & $\ell_{\infty}$ & No  & White & Yes & Both \\
JSMA~\cite{papernot2016limitations} & $\ell_{0}$ & No & White & Yes & Yes \\
C\&W~\cite{carlini2017towards} & $\ell_{0}$, $\ell_{2}$, $\ell_{\infty}$ & No & White & Yes & Yes \\
EAD~\cite{chen2018ead} & $\ell_{1}$, $\ell_{2}$, $\ell_{\infty}$ & No & White & Yes & Yes \\
EOT~\cite{athalye2018synthesizing} & $\ell_{2}$ & Yes & White & Yes & Both \\
BPDA~\cite{athalye2018obfuscated} & $\ell_{2}$, $\ell_{\infty}$ & No & White & Yes & Both \\
OptMargin~\cite{he2018decision} & $\ell_{0}$, $\ell_{2}$, $\ell_{\infty}$ & No & White & Yes & No \\
AutoAttack~\cite{croce2020reliable} & $\ell_{2}$, $\ell_{\infty}$ & No & White & Yes & Both \\
DeepFool~\cite{moosavi2016deepfool} & $\ell_{2}$ & No & White & Yes & No \\
UAP~\cite{moosavi2017universal} & $\ell_{2}$, $\ell_{\infty}$ & No & White & Yes & No \\
UAN~\cite{hayes2018learning} & $\ell_{2}$, $\ell_{\infty}$ & No & White & Yes & Yes \\
ATN~\cite{baluja2018learning} & $\ell_{2}$ & No & White & Yes & Yes \\
FFF~\cite{mopuri2017fast} & $\ell_{\infty}$ & No & White & Yes & No \\
GD-UAP~\cite{mopuri2018generalizable} & $\ell_{\infty}$ & No & White & Yes & No \\
ImageNet-C~\cite{hendrycks2019benchmarking} & $\ell_{\infty}$ & No & Black (Score) & No & No \\
Perlin~\cite{co2019procedural} & $\ell_{\infty}$ & No & Black (Score) & No & No \\
Simplex~\cite{yan2023exploring} & $\ell_{\infty}$ & No & Black (Score) & No & No \\
Worley~\cite{yan2023exploring} & $\ell_{\infty}$ & No & Black (Score) & No & No \\
Papernot et al., 2017~\cite{papernot2017practical} & $\ell_{\infty}$& No & Black (Transfer) & Yes & No\\
Curls\&Whey~\cite{shi2019curls} & $\ell_{2}$& No & Black (Transfer) & Yes & Both \\
Translation-Invariant Attack~\cite{dong2019evading} &  $\ell_{\infty}$& No & Black (Transfer) & Yes & No\\
DI$^2$-FGSM~\cite{xie2019improving} &$\ell_{\infty}$& No & Black (Transfer) & Yes & No\\
VNI-FGSM~\cite{wang2021enhancing} &  $\ell_{\infty}$ & No & Black (Transfer) & Yes & No \\
TREMBA~\cite{huang2019black}& $\ell_{\infty}$ & No & Black (Transfer) & Yes & Both \\
RAP~\cite{qin2022boosting} & $\ell_{2}$, $\ell_{\infty}$ & No & Black (Transfer) & No & Both\\
Yang et al., 2022~\cite{yang2022boosting} &  $\ell_{\infty}$& No & Black (Transfer) & Yes & No\\
NES Attack~\cite{ilyas2018black} & $\ell_{2}$, $\ell_{\infty}$ & No & Black (Score) & No & Both\\
$\mathcal{N}$-Attack~\cite{li2019nattack} & $\ell_{\infty}$ & No & Black (Score) & No & No\\
AdvFlow~\cite{mohaghegh2020advflow} & $\ell_{\infty}$ & No & Black (Score) & No & No\\
ZO-SignSGD~\cite{liu2019signsgd} &  $\ell_{2}$, $\ell_{\infty}$ & No & Black (Score) & No & No\\
Bandit Attack~\cite{ilyas2018prior} & $\ell_{2}$, $\ell_{\infty}$ & No &  Black (Score) & No & Both\\
SimBA~\cite{guo2019simple} & $\ell_{0}$, $\ell_{2}$, $\ell_{\infty}$ & No & Black (Score) & No & Both\\
ECO Attack~\cite{moon2019parsimonious} & $\ell_{2}$, $\ell_{\infty}$ & No & Black (Score) & No & Both\\
Sign Hunter~\cite{al2020sign} & $\ell_{2}$, $\ell_{\infty}$ & No & Black (Score) & No & No\\
Square Attack~\cite{andriushchenko2020square} & $\ell_{2}$, $\ell_{\infty}$ & No & Black (Score) & No & Both\\
$\mathcal{CG-}$ATTACK~\cite{feng2022boosting} & $\ell_{\infty}$ & No & Black (Score) & No & Both\\
Boundary Attack~\cite{brendel2018decision} & $\ell_{2}$& No & Black (Decision) & No & Both\\
OPT~\cite{cheng2019query} & $\ell_{2}$& No & Black (Decision) & No & Both\\
Sign-OPT~\cite{cheng2020sign} & $\ell_{2}$& No & Black (Decision) & No & Both\\
Evolutionary Attack~\cite{dong2019efficient} & $\ell_{2}$& No & Black (Decision) & No & Both\\
CISA~\cite{shi2022query} & $\ell_{2}$& No & Black (Decision) & No & No\\
GeoDA Attack~\cite{rahmati2020geoda} & $\ell_{1}$, $\ell_{2}$, $\ell_{\infty}$& No & Black (Decision) & No & Both\\
HopSkipJumpAttack~\cite{chen2020hopskipjumpattack} & $\ell_{2}$, $\ell_{\infty}$& No & Black (Decision) & No & Both\\
QEBA~\cite{li2020qeba} & $\ell_{2}$, $\ell_{\infty}$& No & Black (Decision) & No & Both\\
Sign Flip Attack~\cite{chen2020boosting} & $\ell_{\infty}$& No & Black (Decision) & No & Both\\
RayS~\cite{chen2020rays} & $\ell_{\infty}$& No & Black (Decision) & No & No\\
\bottomrule
\end{tabularx}
\end{table}
\subsection{White-box Attacks}
\indent
\par The first milestone work of adversarial attack is Limited Memory Broyden-Fletcher-Goldfarb-Shanno (L-BFGS) attack~\cite{szegedy2014intriguing}, which aims at finding an imperceptible minimum input perturbation in the constraint space of inputs. Eq. (\ref{eq:lbfgs}) defines an L-BFGS attack as a box-constrained optimization \textcolor{black}{for the maximization of the loss and the minimization of the perturbation norm}:
\begin{equation}\label{eq:lbfgs}
\min _{\delta} c\|\delta\|_{2}+\operatorname{Loss}(x+\delta, l),
\end{equation}
where the symbols $x$, $\delta$, and $l$ denote the data, the perturbation, and the label. The adversaries perform the line-search mechanism to \textcolor{black}{find the minimum tradeoff parameter $c$}. Goodfellow et al.~\cite{goodfellow2015explaining} propose an one-step iterative gradient-based attack method defined in Eq. (\ref{eq:fgsm}). Basic Iterative Method (BIM or I-FGSM)~\cite{kurakin2016adversarial} iteratively solves $\delta$ and updates new adversarial samples based on FGSM~\cite{goodfellow2015explaining} in multiple steps defined in Eq. (\ref{eq:pgd}). The PGD method~\cite{madry2018towards} will project the perturbed input back onto the norm ball. The I-FGSM method combined with the momentum method can evolve the momentum I-FGSM method (MIM)~\cite{dong2018boosting}. Eq. (\ref{eq:mim}) defines the procedure of the MIM attack, where $\nabla_{i} J(\cdot)$ is the gradient of the specific time step.
\begin{equation}\label{eq:mim}
\begin{aligned} x_{i+1} & =\operatorname{Clip}\left\{x_{i}+\varepsilon \cdot \frac{\nabla_{i+1} J(\cdot)}{\left\|\nabla_{i+1} J(\cdot)\right\|_{2}}\right\} \\ \nabla_{i+1} J(\cdot) & =\mu \cdot \nabla_{i} J(\cdot)+\frac{\nabla_{x} \operatorname{Loss}\left(x_{i}, y\right)}{\left\|\nabla_{x} \operatorname{Loss}\left(x_{i}, y\right)\right\|_{1}}\end{aligned}.
\end{equation}
The Jacobian Salient Map Attack (JSMA) method~\cite{papernot2016limitations} can use salient map defined in Eq. (\ref{eq:jsma}) learned by the neural networks to generate the adversarial examples.
\begin{equation}\label{eq:jsma}
S(\mathbf{X}, t)[i]=\left\{\begin{array}{l}0 \text { if } \frac{\partial \mathbf{F}_{t}(\mathbf{X})}{\partial \mathbf{X}_{i}}<0 \text { or } \sum_{j \neq t} \frac{\partial \mathbf{F}_{j}(\mathbf{X})}{\partial \mathbf{X}_{i}}>0 \\ \left(\frac{\partial \mathbf{F}_{t}(\mathbf{X})}{\partial \mathbf{X}_{i}}\right)\left|\sum_{j \neq t} \frac{\partial \mathbf{F}_{j}(\mathbf{X})}{\partial \mathbf{X}_{i}}\right| \text { otherwise }\end{array}\right.,
\end{equation}
where $\mathbf{F(.)}$ is the salient map denoted by Jacobian matrix, and $S(\mathbf{X}, t)[i]$ is a corresponding saliency map. The adversary modifies the input feature with the saliency map to realize the deceit.
\par C\&W attack~\cite{carlini2017towards} tries to find small $\delta$ in $\ell_{0}$, $\ell_{2}$, and $\ell_{\infty}$ norm. It is an adaptive attack method that the attacker has knowledge of the defense strategies and specifically designs an attack to circumvent or disrupt these defenses. The adaptive attack method is dynamic and purposeful, meaning it adapts to the specific defense mechanisms of the target model. Eq. (\ref{eq:cw}) defines the optimization scheme of the C\&W attack.
\begin{equation}\label{eq:cw}
\begin{array}{c}\min _{\delta}\|\delta\|_{p}+c \cdot f \mid(x+\delta) \\ f(x+\delta)=\max \left(\max \left\{Z(x+\delta)_{i}: i \neq t\right\}-Z(x+\delta)_{t},-\mathcal{K}\right)\end{array},
\end{equation}
where $c$ is a hyperparameter, $f(·)$ is an artificially defined function, and $\mathcal{K}$ is the constraint to assist the generation of adversarial examples. After the proposal of the C\&W attack, other variant methods of adaptive attacks have been proposed. The EAD method (Elastic-net Attacks to Deep Neural Networks)~\cite{chen2018ead} transforms the process of attacking Deep Neural Networks (DNNs) using adversarial samples into an optimization problem using elastic-regularized net. OptMargin~\cite{he2018decision} is another extension of the C\&W attack by replacing one objective function with multiple objective functions around the data $x$. The EOT (Expectation Over Transformation) method is a generalized framework that allows for the construction of adversarial examples that remains the deceptive effect on selected transformation distributions $\mathcal{T}$. The core idea is to constrain the distance between the adversarial input and the original input in the optimization process. Eq. (\ref{eq:eot_perturbation}) defines the defined perturbation of the EOT method, and Eq. (\ref{eq:eot_optimization}) describes the formulation of the optimization problem.
\begin{equation}\label{eq:eot_perturbation}
\delta=\mathbb{E}_{t \sim \mathcal{T}}\left[d\left(t\left(x^{\prime}\right), t(x)\right)\right],
\end{equation}
\begin{equation}\label{eq:eot_optimization}
\begin{array}{cl}\underset{x^{\prime}}{\arg \max } & \mathbb{E}_{t \sim \mathcal{T}}\left[\log P\left(y_{t} \mid t\left(x^{\prime}\right)\right)\right] \\ \text { subject to } & \mathbb{E}_{t \sim \mathcal{T}}\left[d\left(t\left(x^{\prime}\right), t(x)\right)\right]<\epsilon \\ & x \in[0,1]^{d}\end{array},
\end{equation}
where $\mathcal{T}$ is the distribution and $t(.)$ means the transformation which is robust to noise, distortion, and affine transformations. In the landmark work on robustness~\cite{athalye2018obfuscated}, Athalye et al. identified the phenomenon of obfuscated gradients, highlighting false security in certain defense methods under iterative optimization attacks. They also discovered three types of gradient phenomena leading to confusion: Shattered Gradients, Stochastic Gradients, and Vanishing/Exploding Gradients. Moreover, they proposed three attack techniques for inspecting obfuscated gradient types: Backward Pass Differentiable Approximation (BPDA), Expectation Over Transformation (EOT), and Reparameterization~\cite{athalye2018obfuscated}. Another notable work is the AutoAttack method~\cite{croce2020reliable}, which aims to address the misleading impression of robustness by identifying evaluation pitfalls. Based on an open leaderboard, the AutoAttack method can evaluate defense methods for potential gradient obfuscation or masking. It has now become an unwritten rule that defense methods should be assessed on the AutoAttack benchmark.
\subsection{Universal Adversarial Perturbations}
\indent
\par The aforementioned white-box attack methodologies are tailored to specific models, thereby catalyzing inquiries into model-agnostic assault techniques. Moosavi-Dezfooli et al.~\cite{moosavi2016deepfool} computes the minimal adversarial disturbance necessary for a more accurate assessment of robustness. This development has spurred further investigation into universal adversarial perturbations (UAP), capable of affecting a wide range of model architectures. The UAP method~\cite{moosavi2017universal} identifies perturbations that exhibit transferability across diverse models. Other generative methods including Universal Adversarial Networks (UAN)~\cite{hayes2018learning} and Adversarial Transform Networks (ATN)~\cite{hayes2018learning} can generate data-specific universal perturbations, while the Fast Feature Fool method (FFF)~\cite{mopuri2017fast} and Generalizable data-free UAP (GD-UAP)~\cite{mopuri2018generalizable} can generate data-independent universal perturbations. In contrast to white-box universal adversarial perturbations, black-box counterparts offer broader applicability in real-world contexts, transforming external security threats into metrics for evaluating internal system safety. ImageNet-C~\cite{hendrycks2019benchmarking} is a benchmark to evaluate the robustness of neural networks to common perturbations. The corruptions include Gaussian noise, shot noise, impulse noise, defocus blur, glass blur, motion blur, zoom blur, snow, frost, fog, brightness, contrast, elastic transform, pixelate, jpeg compression, speckle noise, Gaussian blur, spatter, and saturation. The utilization of procedural noise functions in computer graphics~\cite{co2019procedural,yan2023exploring} can also generate the textures to deceive the neural networks. The application of ImageNet-C~\cite{hendrycks2019benchmarking}, Perlin Noise~\cite{co2019procedural}, Simplex Noise~\cite{yan2023exploring}, Worley Noise~\cite{yan2023exploring}, and other black-box UAPs can help simulate the adverse weathers and sensor disturbances in the virtual experiment. Such plug-ins are significant for achieving SOTIF in automated driving.
\subsection{Black-box Attacks}
\indent
\par Launching the white-box attacks requires knowledge of model structures or data distributions, which limits its scope in the actual applications. In most cases, attackers and defenders do not know each other. Therefore, it is significant to study adversarial robustness under black-box attacks. This sub-section will give a review of the classical black-box adversarial attack methods.
\subsubsection{Transfer-based Black-box Attacks}
\indent
\par A promising avenue for black-box attack methodologies involves the deployment of transfer-based attacks. Within this framework, adversaries cultivate a surrogate model to mimic the targeted system. It is achieved by employing inputs artificially crafted by the adversary, which are subsequently classified by the target model to predict the wrong labels. Upon mounting a successful attack on the surrogate model, the adversary is then capable of extrapolating the malevolent data to the target model to launch the attack.
\par In the past decade, many significant studies related to transfer-based black-box attacks have been proposed. Papernot et al.~\cite{papernot2017practical} propose the first pioneering work of transfer-based adversarial attack. In this milestone work~\cite{papernot2017practical}, the attacker generates the adversarial synthetic inputs by a Jacobian-based heuristic and crafts the adversarial examples with a high attack success rate to invade the MLaaS systems. Shi et al.~\cite{shi2019curls} propose a Curls \& Whey optimization mechanism to boost the transfer-based attack that the adversaries ``curl" up the iterative invasion trajectories to add more diversities and transferabilities in the malicious outputs and further squeeze the ``whey" of noise to boost the robustness of perturbations. The white-box adversarial examples would usually be correlated with the discriminative regions of models or gradient trajectories in the optimization process, leading to difficulties in the transferability of adversarial attacks. The tranferability can be improved by data augmentation, including MIM~\cite{dong2018boosting} and Diverse Inputs Iterative Fast Gradient Sign Method (DI$^2$-FGSM)~\cite{xie2019improving}. Wang et al.~\cite{wang2021enhancing} propose a variant tuning momentum iterative FGSM method (VNI-FGSM) to boost the attack performance. Huang et al.~\cite{huang2019black} proposed an attack method based on transferable model-based embedding called TRansferable EMbedding-based Black-box Attack (TREMBA). This approach utilizes pre-trained models to learn a low-dimensional embedding space and search within the space to generate adversarial perturbations with high-level semantic patterns to improve the effectiveness of black-box attacks. Yang et al.~\cite{yang2022boosting} propose a method to attack the target model via the hierarchical generative networks. Qin et al.~\cite{qin2022boosting} propose a method to achieve both targeted and untargeted attacks via the Reverse Adversarial Perturbation (RAP), which finds the stable adversarial examples by minimizing the maximum loss value within a local neighborhood.
\subsubsection{Score-based Black-box Attacks}
\indent
\par Within a black-box query-based adversarial attack, attackers lack internal model details like weights and structure. Instead, they utilize input and output information to craft effective adversarial examples. One popular methodology is the score-based attack: the attacker adjusts their strategy using the model's score/probability output. The attack typically involves: 1. The attacker makes an exploratory query; 2. The model returns a confidence score; 3. The methods like zeroth-order optimization~\cite{liu2020primer} generate adversarial examples, potentially tricking the model.
\par The Natural Evolution Strategy (NES) attack methodology, introduced by Ilyas et al.~\cite{ilyas2018black}, represents a foundational approach to score-based black-box attacks. This pioneering work delineates three distinct real-world scenarios: the query-limited setting, the partial-information setting, and the label-only setting. The NES method adeptly generates black-box adversarial examples within a query-limited context by estimating gradients and constructing adversarial examples through the application of the PGD algorithm on the estimated gradients. In scenarios characterized by partial information, NES strategically perturbs the image by projecting it onto a sphere centered around the original image, thereby maximizing the likelihood of misclassification into the target category while ensuring inclusion within the top $k$ predicted classes. However, the long query time is an obstacle that limits the scalability of the NES method.
\par Many other studies enhance the paradigm started from the NES attack. Li et al.~\cite{li2019nattack} propose a method called $\mathcal{N}$-Attack to find a probability density distribution in a narrow region centered on the input, from which sampling can increase the success of a black-box attack. AdvFlow~\cite{mohaghegh2020advflow} is an extension of $\mathcal{N}$-Attack in which the adversary exploits the normalizing flows for constructing the probability density function of adversarial examples. Liu et al.~\cite{liu2019signsgd} design a zeroth-order stochastic optimization algorithm (ZO-signSGD), which employs the dual advantages of gradient-free operations and the signSGD mechanism to address the problem of black-box attacks. Ilyas et al.~\cite{ilyas2018prior} form the construction of such a black-box attack as a gradient estimation problem and prove that a least-square estimator is a feasible way to solve this problem. They propose a method based on bandit optimization, enabling the adversaries to integrate priors into the attack settings. Guo et al.~\cite{guo2019simple} propose a simple black-box attack method (SimBA) that utilizes a finite assumption of continuous-valued confidence scores to construct the adversarial images by randomly selecting orthogonal basis vectors and adding or subtracting them in the manipulation process. To augment the efficacy of score-based black-box attacks, Moon et al.~\cite{moon2019parsimonious} propose an efficient combinatorial optimization (ECO) attack method to generate the adversarial perturbations. Further contributing to advancements in this field, Dujaili et al.~\cite{al2020sign} propose the SignHunter algorithm, which innovatively estimates the sign bit of the gradient during black-box attacks. By leveraging the divisibility characteristic of directional derivatives in the loss function related to the attack, SignHunter employs a partitioning strategy coupled with adaptive querying to ascertain the gradient's sign bit. This method stands out for its remarkable accuracy and efficiency, significantly improving existing techniques. Andriushchenko et al.~\cite{andriushchenko2020square} propose the Square Attack method that utilizes a randomized search scheme, ensuring that the perturbation is strategically crafted near the feasible set of the boundary at each iteration. Applying this black-box method in the object detection task is also successful~\cite{liang2021parallel}.
\par Two main challenges remain for the research of the score-based black-box attacks. First, the efficiency problem inherited from the milestone NES attack~\cite{ilyas2018black} is still an open issue. Second, some studies~\cite{yan2023wavelet,zheng2023blackboxbench} demonstrate that these classical black-box methods have difficulties attacking the relatively robust structure like WideResNet~\cite{zagoruyko2016wide}, Vision Transformer (ViT)~\cite{dosovitskiy2021image}, and SwinTransformer~\cite{liu2021swin}. The novel score-based black-box attack research in the new paradigm is meaningful.
\subsubsection{Decision-based Black-box Attacks}
\indent
\par In many real-world applications, the confidence scores of neural network outputs are invisible to users. Instead, the MLaaS systems provide a final decision, not uncertainty information. It emphasizes the importance of decision-based black-box attacks. These are applicable to real AI systems like intelligent vehicles. The decision-based attacks require less knowledge than the transfer-based attacks and are more difficult to defend against than score-based attacks.
\par The Boundary Attack~\cite{brendel2018decision} is the first decision-based attack method in which the adversary starts from a large adversarial perturbation and then seeks to reduce the perturbation while staying the beguiling effect. Cheng et al.~\cite{cheng2019query} postulate that the random-walk method on the boundary, which requires many queries, lacks convergence guarantees. Based on the zeroth order optimization, the OPT method~\cite{cheng2019query} addresses the issue where the decision-based black-box attack is formulated as a real-valued continuous optimization problem. The extension work of Sign-OPT incorporates a direct estimation of the sign of gradient at any direction to the OPT framework~\cite{cheng2020sign}. Another extension of Boundary Attack~\cite{brendel2018decision} is Customized Iteration and Sampling Attack (CISA)~\cite{shi2022query} that the adversary estimates the distance based on a dual-direction iterative trajectory from the nearby decision boundary for iterative search of adversarial examples. The Evolutionary Attack~\cite{dong2019efficient} method models the local geometry in the search direction and reduces the dimension of the sampling space of adversarial examples. Rahmati et al.~\cite{rahmati2020geoda} propose a geometry-based framework named Geometric Decision-based Attack (GeoDA) to generate black-box adversarial samples where each query returns the highest confidence label of the classifier. The GeoDA framework builds on the assumption that the decision boundaries of neural networks typically have small mean curvature observations in the neighborhood of the data sample. The authors propose an efficient iterative algorithm for generating black-box perturbations with a small $p$-paradigm ($p \geq 1$), which is validated by the attack experiments on state-of-the-art image classifiers. Chen et al.~\cite{chen2020hopskipjumpattack} propose a novel HopSkipJumpAttack method that generates adversarial samples with the Monte Carlo estimation method in a hard-label setting. The algorithm is based on a new gradient direction estimation that uses binary information to estimate the gradient direction on the decision boundary and approximates the optimal solution iteratively. Implementing the zeroth-order gradient estimation in the low-dimensional subspace instead of the original space is a potential query-efficient boundary-based black-box attack (QEBA) method~\cite{li2020qeba}. Chen et al.~\cite{chen2020boosting} show that randomly flipping the signs of the entries improves the effectiveness and efficiency of the adversarial attack process. The Ray Searching (RayS) method~\cite{chen2020rays} addresses the inefficiency of decision-based black-box attacks. It builds on the discrete modeling of continuous problems to avoid gradient estimation. Moreover, it eliminates all unnecessary searches through a quick checking step that surprisingly reduces the number of queries required for the attack.
\subsection{Physical Attacks}
\begin{table}[!t]
\scriptsize
\centering
\caption{Classical physical attack methods in automated driving. The main text gives the full names of abbreviations.}
\label{tab:phy_attack_method}
\begin{tabularx}{\textwidth}{@{} l @{\hspace{8pt}} *{8}{X} @{}}
\toprule
Method & Knowledge & Tasks \\
\midrule
RP$_{2}$~\cite{eykholt2018robust} & White & Traffic sign recognition \\
RP$_{2}$D~\cite{song2018physical} & White & Traffic sign detection \\
CAMOU~\cite{zhang2019camou} & Black & Vehicle detection\\
Adversarial camera sticker~\cite{li2019adversarial}  & White & Traffic sign recognition\\ 
FIR~\cite{zhao2019seeing}  & White & Traffic sign detection \\
ERG~\cite{zhao2019seeing}  & White & Traffic sign detection \\
AdvCam~\cite{duan2020adversarial} & White & Traffic sign recognition \\
PhysGAN~\cite{kong2020physgan} & White & Traffic sign recognition \\
ER~\cite{wu2020physical} & Black & Vehicle detection \\
UPC~\cite{huang2020universal} & Black & Vehicle detection \\
Wu et al.~\cite{wu2020making} & White & Person detection \\
Xu et al.~\cite{xu2020adversarial} & White & Person detection \\
Boloor et al.~\cite{boloor2020attacking} & Black & Road line segmentation \\
Yamanaka et al.~\cite{yamanaka2020adversarial} & White & Monocular depth estimation \\
Sun et al.~\cite{sun2020towards} & Black & LiDAR perception \\
Tu et al.~\cite{tu2020physically} & White & LiDAR perception \\
IAP~\cite{bai2021inconspicuous} & Black & Traffic sign recognition \\
Adversarial Patch~\cite{ye2021patch} & White & Traffic sign recognition \\
SLAP~\cite{lovisotto2021slap} & Black & Traffic sign recognition \\
Adversarial Laser~\cite{duan2021adversarial} & Black & Traffic sign recognition \\
Rolling Shutter Effect Attack~\cite{sayles2021invisible} & Black & Traffic sign recognition \\
DAS~\cite{wang2021dual}  & White & Vehicle detection \\
Zolfi et al.~\cite{zolfi2021translucent} & White & Traffic light detection \\
Sato et al.~\cite{sato2021dirty} & White & Road line segmentation \\
Cao et al.~\cite{cao2021invisible} & White & Multi-sensor fusion \\
Shadow Attack~\cite{zhong2022shadows} & Black & Traffic sign recognition \\
DTA~\cite{suryanto2022dta} & White & Vehicle detection \\
Cheng et al.~\cite{cheng2022physical} & White & Monocular depth estimation\\
Dos Attack~\cite{wan2022too} & White & Navigation and planning\\
RP$_{2}$-CAM~\cite{yan2023adversarial} & White & Traffic sign recognition \\
Cao et al.~\cite{cao2023you} & White & LiDAR perception\\
\bottomrule
\end{tabularx}
\end{table}
\indent
\par In real-world applications, physical attacks may have more significant impact and research value than digital attacks, especially in the context of autonomous driving. Physical attacks typically encompass three stages: 1) the generation of adversarial perturbations in digital space; 2) the transformation of digital perturbations into physical perturbations with robustness guarantees; and 3) the evaluation of physical perturbations using scanners, cameras, or LiDAR devices. Table \ref{tab:phy_attack_method} lists well-established physical adversarial attack methods, which will be further described in subsequent sections.
\par Physical attacks should ensure two types of robustness. First, the robustness of digital-to-physical transformation: color space sensitivity can cause physical attack instability. The non-printability score (NPS) metric helps address this issue \cite{sharif2016accessorize}. Second, the robustness of physical-to-digital transformation: physical adversarial examples should maintain deception under camera distortion, spectral interference, and incomplete echoes in LiDAR and Radar. The adaptive EOT (Expectation Over Transformation) attack method \cite{athalye2018synthesizing} can increase adversarial example robustness across scale or rotation changes. Most white-box physical adversarial attacks follow the NPS-EOT combination paradigm.
\par Several vision tasks in automated driving would be disturbed by the physical adversarial examples, including traffic sign recognition and detection, traffic recognition, vehicle detection, road line segmentation, monocular depth estimation, and LiDAR perception. 
\subsubsection{Traffic Sign Recognition and Detection}
\indent
\par Eykholt et al.~\cite{eykholt2018robust} propose a general attack algorithm, Robust Physical Perturbations (RP$_2$), to generate robust visual adversarial perturbations under different physical conditions. This method can attack the ``STOP" sign as the speed-limit sign. This attack paradigm can also be applied in the object detection task to fool the state-of-the-art (SOTA) model~\cite{song2018physical}. Li et al.~\cite{li2019adversarial} propose a novel method in which the adversaries manipulate the translucent sticker over the lens of a camera to fool the traffic sign classifier. Zhao et al.~\cite{zhao2019seeing} attempt to attack the feature extraction process to boost the physical attack performance. In addressing the Hiding Attack (HA) scenario, they introduce the feature-interference reinforcement (FIR) method alongside the enhanced realistic constraints generation (ERG) approach to bolster robustness. Conversely, for the Appearing Attack (AA), they devise the nested-AE framework, which integrates two autoencoders (AEs) to compromise object detectors effectively at both long and short distances. The patch-based physical attack is easily identified by the human observer, which remains a massive challenge in security research. The Adversarial Camouflage (AdvCam) method~\cite{duan2020adversarial} is proposed to incorporate the natural style in the physical adversarial examples so that the crypticity of the adversarial examples is increased. Furthermore, the generative adversarial networks (GAN)~\cite{goodfellow2014generative} can be utilized as a data augmentation method to enhance the adversarial attack~\cite{kong2020physgan}. Ye et al.~\cite{ye2021patch} apply the adversarial patch method~\cite{brown2017adversarial} in the attack on the traffic sign recognition. Bai et al.~\cite{bai2021inconspicuous} attempt to generate the inconspicuous adversarial patches (IAP) to boost the transferability. The IAP method uses the patch generation process in a coarse-to-fine way by utilizing multiple-scale generative models. Lovisotto et al.~\cite{lovisotto2021slap} use a light projector to craft the attacks with the generated Short-Lived adversarial perturbations (SLAP). The laser jamming~\cite{duan2021adversarial}, rolling shutter effect~\cite{sayles2021invisible}, and even shadows~\cite{zhong2022shadows} can be utilized to craft the adversarial examples to fool the traffic sign classifiers. Adversarial vulnerability can be regarded as the causal confounding effect. Therefore, Yan et al.~\cite{yan2023adversarial} attack the traffic sign with the guidance of class activation map (CAM)~\cite{zhou2016learning} to find the sensitive regions of the targeted attack class. The recent study~\cite{yan2023adversarial} finds that the attack difficulties increase after the model structures evolved from Convolution Neural Networks (CNNs)~\cite{szegedy2015going,he2016deep,zagoruyko2016wide} to ViTs~\cite{dosovitskiy2021image,liu2021swin}, which raises a new research focus.
\subsubsection{Vehicle Detection}
\indent
\par The research on attacks on the vehicle detection model has a huge impact on military applications. Zhang et al.~\cite{zhang2019camou} propose the first physical vehicle camouflage inspired by both the research of adversarial examples~\cite{goodfellow2015explaining} and GANs~\cite{goodfellow2014generative}. Such a milestone method implements a camouflage pattern to hide the vehicle from being detected by state-of-the-art CNN-based object detectors~\cite{he2017mask,redmon2018yolov3}. The proposed method alternates between two threads. First, the attacker trains a neural approximation function to simulate how the simulator applies camouflage to the vehicle and how the vehicle detector performs, given an image of the camouflaged vehicle. Second, the attacker can minimize the approximation detection score by searching for the optimal camouflage. Wu et al.~\cite{wu2020physical} propose an Enlarge-and-Repeat process (ER) method and a Discrete Searching method to generate the adversarial examples fooling the vehicle detectors. The methods effectively produce the mosaic-like adversarial vehicle textures without using the detector's model weights and differential rendering procedure. The limitation of the method is that it is simulated only in the Carla software. Huang et al.~\cite{huang2020universal} propose a Universal Physical Camouflage (UPC) Attack that can fool the region regressors and classifiers simultaneously. Wang et al.~\cite{wang2021dual} propose a Dual Attention Suppression (DAS) method to inhibit model and human attention. Suryanto et al.~\cite{suryanto2022dta} propose a Differentiable Transformation Attack (DTA) method that the adversary utilizes a Differentiable Transformation Network (DTN) to learn the expected transformations of rendered objects and generate a robust camouflage texture to attack the vehicle detectors with a wide range of transformations.
\subsubsection{Person Detection}
\indent
\par Another potential security threat in automated driving is the assault aimed at the person detectors. The attackers can craft the adversarial T-shirts, which are robust under different motion gestures~\cite{wu2020making,xu2020adversarial}. This malicious clothing can play tricks on both surveillance systems and detectors on pedestrians in automated driving. These pieces of research would also be important in the military self-driving applications.
\subsubsection{Road Line Segmentation}
\indent
\par In automated driving, road line segmentation is a vital task to ensure the vehicles are in the right line locations. However, it is vulnerable to adversarial attacks. Boloor et al.~\cite{boloor2020attacking} propose a query-based attack that produces a black malicious line to fool the neural networks. To address the problem of camera frame inter-dependencies influenced by vehicle control, Sato et al.~\cite{sato2021dirty} formulate the problem with a security-critical attack goal and propose a novel attack method based on the dirty road patches.
\subsubsection{Monocular Depth Estimation}
\indent
\par Monocular depth estimation refers to estimating the depth information of each pixel in a scene from a single image. It usually involves using computer vision techniques and algorithms to analyze information such as texture, occlusion relationships, and perspective transformations in the image to infer spatial relationships and distances between pixels. It is also sensitive to adversarial attacks. Yamanaka et al.~\cite{yamanaka2020adversarial} first apply the patch-based attack method in the monocular depth estimation task. Cheng et al.~\cite{cheng2022physical} attack the depth estimation model by generating covert object-oriented adversarial patches, and the proposed attack procedure searches the optimization region as well as utilizes the symmetrization methods to deal with the overall contour region to find the most effective attack method. The attack method can attack different target objects and models in real driving scenarios, leading to depth estimation errors and decreased object detection success.
\subsubsection{LiDAR Perception and Multi-sensor Fusion}
\indent
\par LiDAR sensing plays a vital role in automated driving, which utilizes laser radar (LiDAR) technology to understand the surrounding environment and obtain road information. The LiDAR device can measure the distance from the body to an obstacle by emitting laser light outward. When it encounters an object, the laser light is reflected and received by a complementary metal–oxide semiconductor (CMOS) sensor reflects and receives the laser light. By combining the real-time Global Positioning System (GPS), inertial navigation information, and the calculation of the emission angle, the system can derive the coordinate orientation and distance information of the object in front. The LiDAR device, with powerful information perception and processing capabilities, can sense the road environment and control the vehicle to achieve the intended goal. Sun et al.~\cite{sun2020towards} conduct the first study on adversarial examples of LiDAR perception in automated driving to explore general vulnerabilities in current LiDAR-based perception architectures and find that neglected occlusion patterns in LiDAR point clouds make self-driving cars vulnerable to spoofing attacks. Sun et al.~\cite{sun2020towards} construct the first black-box spoofing attack based and successfully attack the PointPillars model~\cite{lang2019pointpillars} and the PointRCNN model~\cite{shi2019pointrcnn}. However, the proposed method of Sun et al.~\cite{sun2020towards} named LidarAdv only considers the specific frame. Tu et al.~\cite{tu2020physically} propose a method to generate the adversarial mesh which can be placed on a vehicle roof to hide the malicious object and implement the defense experiment under the attacks with the method of data augmentation and Fast Adversarial Training~\cite{wong2020fast}. Cao et al.~\cite{cao2023you} propose a novel attack method called Physical Removal Attack (PRA), which is capable of selectively removing LiDAR point cloud data of real obstacles by utilizing laser jamming technology, thus causing the obstacle detector of self-driving cars to fail to recognize and locate obstacles, which in turn enables the cars to make dangerous self-driving decisions. Cao et al.~\cite{cao2021invisible} design an attack pipeline with non-differentiable cell-level aggregated features to fool both cameras and LiDAR devices with the invisible perturbations. Recently, Wan et al.~\cite{wan2022too} have investigated a Semantic DoS (Semantic Denial of Service) vulnerability in self-driving planning systems that could lead to unexpected decision-making behaviors in self-driving vehicles, such as sudden braking or abandoning lane changes. This research designs a vulnerability discovery system called PlanFuzz and demonstrates the severity of the vulnerability and possible exploits through case studies of three attack scenarios.
\subsubsection{Perspectives of Physical Attacks}
\indent
\par Intelligent connected vehicles (ICV) have achieved mass production in the past few years~\cite{kuang2018intelligent}. Many companies have put their self-driving vehicles into the real-world road testing phases. In the research field, the performance metrics on the KITTI dataset~\cite{geiger2012we} tend to be saturated. On the other hand, the neural network models are vulnerable to the attacks. Compared to the digital adversarial examples, the physical adversarial examples would link more to cybersecurity, which is more severe in real applications. The research on physical attacks and their associated defense strategies is still crucial in the future.
\section{Defense}
 \begin{figure}[!t]
        \centering
\includegraphics[width=0.85\hsize]{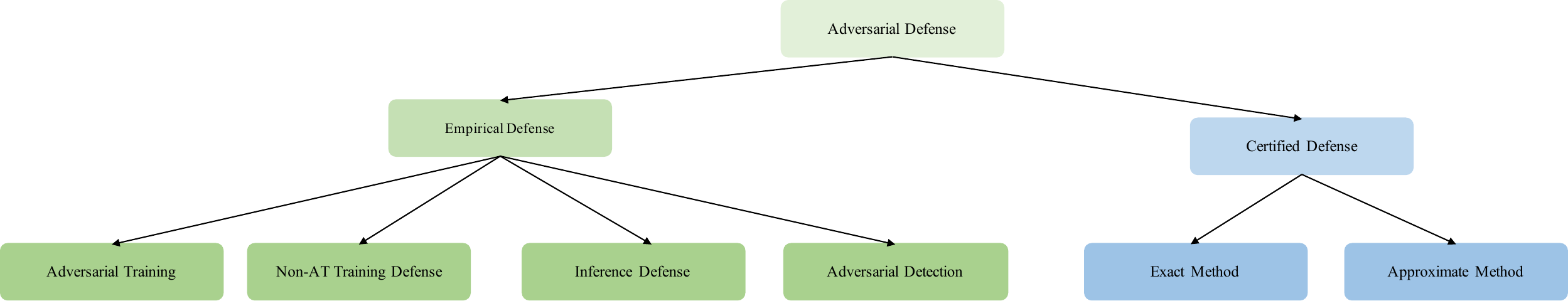}
\caption{The category of mainstream adversarial defense method.}
\label{fig:defense_category}
\end{figure}
 \begin{figure}[!t]
        \centering
\includegraphics[width=0.85\hsize]{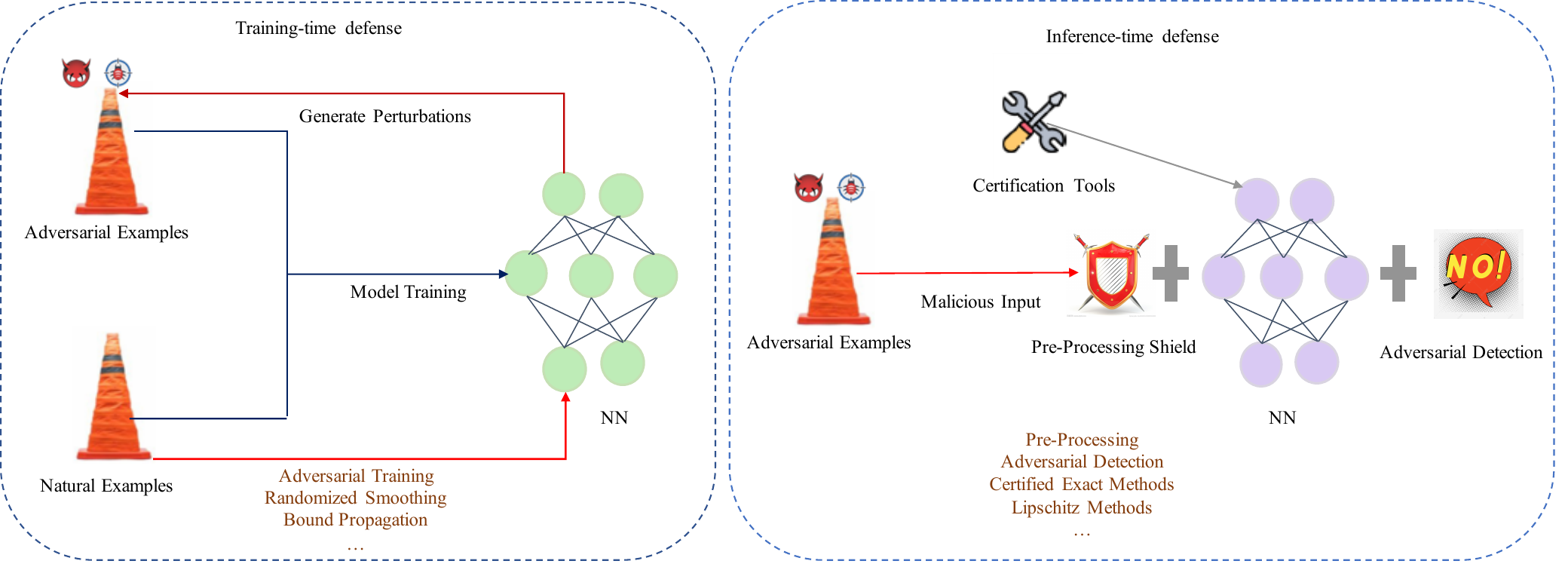}
\caption{Different deployment stages of defense methods.}
\label{fig:defense_training_inference}
\end{figure}
\begin{table}[!t]
\scriptsize
\centering
\caption{\textcolor{black}{Classical empirical defense methods categorized by addressed problems. The main text provides the full names of abbreviations.}}
\label{tab:empirical_defense_methods}

\textbf{Robust Generalization}
\begin{tabularx}{\textwidth}{@{} l @{\hspace{5pt}} *{3}{X} @{}}
\toprule
\textbf{Method} & \textbf{Defending Attack Types} & \textbf{Adversarial Training?} \\
\midrule
Defense Distillation~\cite{papernot2016distillation} & White & No \\
Vanilla Adversarial Training~\cite{madry2018towards} & White & Yes \\
Ensemble Adversarial Training~\cite{tramer2018ensemble} & White & Yes \\
Deep Defense~\cite{yan2018deep} & White & No \\
ATDA~\cite{song2019improving} & White & Yes \\
TRADES~\cite{zhang2019theoretically} & White & Yes \\
MART~\cite{wang2020improving} & White & Yes \\
CCAT~\cite{stutz2020confidence} & White & Yes \\
Friendly Adversarial Training~\cite{zhang2020attacks} & White & Yes \\
DVERGE~\cite{yang2020dverge} & White & Yes \\
Bag of Tricks for AT~\cite{pang2021bag} & White & Yes \\
Adversarial Wavelet Training~\cite{yan2023wavelet} & White & Yes \\
DM-Improves-AT~\cite{wang2023better} & White & Yes \\
GeodesicAT~\cite{yan2024enhance} & White & Yes \\
\bottomrule
\end{tabularx}

\vspace{10pt}

\textbf{Adversarial Detection}
\begin{tabularx}{\textwidth}{@{} l @{\hspace{5pt}} *{3}{X} @{}}
\toprule
\textbf{Method} & \textbf{Defending Attack Types} & \textbf{Adversarial Training?} \\
\midrule
Metzen et al.~\cite{metzen2016detecting} & White & No \\
SafetyNet~\cite{lu2017safetynet} & White & No \\
MagNet~\cite{meng2017magnet} & White & No \\
GMM~\cite{zheng2018robust} & White & No \\
Mahalanobis distance~\cite{lee2018simple} & White & No \\
Reverse Cross Entropy~\cite{pang2018towards} & White & No \\
CD-VAE~\cite{yang2021class} & White & No \\
Libre~\cite{deng2021libre} & White & No \\
Blacklight~\cite{li2022blacklight} & Black & No \\
PRADA~\cite{juuti2019prada} & Black & No \\
SD~\cite{chen2020stateful} & Black & No \\
\bottomrule
\end{tabularx}

\vspace{10pt}

\textbf{Inference-time Defense}
\begin{tabularx}{\textwidth}{@{} l @{\hspace{5pt}} *{3}{X} @{}}
\toprule
\textbf{Method} & \textbf{Defending Attack Types} & \textbf{Adversarial Training?} \\
\midrule
Input Transformations~\cite{guo2018countering} & White & No \\
PixelDefend~\cite{song2018pixeldefend} & White & No \\
Randomization~\cite{xie2018mitigating} & White & No \\
BaRT~\cite{raff2019barrage} & White & No \\
Mixup Inference~\cite{pang2020mixup} & White & No \\
RND~\cite{qin2021random} & Black & No \\
DiffPure~\cite{nie2022diffusion} & White & No \\
AAA~\cite{chen2022adversarial} & Black & No \\
Boundary defense~\cite{aithal2022boundary} & Black & No \\
Anti-adversaries~\cite{alfarra2022combating} & White & No \\
Dent~\cite{wang2021fighting} & Black & No \\
EBM+DSM~\cite{yoon2021adversarial} & White & No \\
SOAP~\cite{shi2021online} & White & No \\
\bottomrule
\end{tabularx}

\vspace{10pt}

\textbf{Training Efficiency}
\begin{tabularx}{\textwidth}{@{} l @{\hspace{5pt}} *{3}{X} @{}}
\toprule
\textbf{Method} & \textbf{Defending Attack Types} & \textbf{Adversarial Training?} \\
\midrule
Free Adversarial Training~\cite{shafahi2019adversarial} & White & Yes \\
YOPO~\cite{zhang2019you} & White & Yes \\
Fast Adversarial Training~\cite{wong2020fast} & White & Yes \\
Local Linearity Regularizer~\cite{qin2019adversarial} & White & Yes \\
GradAlign~\cite{andriushchenko2020understanding} & White & Yes \\
FrequencyLowCut Pooling~\cite{grabinski2022frequencylowcut} & White & Yes \\
Robust critical fine-tuning~\cite{zhu2023improving} & White & Yes \\
\bottomrule
\end{tabularx}

\end{table}

\indent
\par This section reviews the classical defense methods proposed in the past few years. One type of defense is empirical defense, which relies heavily on practical experience and intuitive judgment to make a defense against a specific attack. The other type is certified defense, which does not care about the type of adversarial noise but constructs a strict robustness tight lower bound through mathematical or physical modeling. Most defense methods can be categorized under these two categories. Figure \ref{fig:defense_category} describes a coarse-grained category of current defense methods. The empirical defense methods include adversarial training (AT), non-AT training defense, inference defense, and adversarial detection. The certified defense methods can be classified as exact methods or approximate methods. Fig. \ref{fig:defense_training_inference} illustrates the different deployment stages of defense methods. In the training stage, adversarial training and other approximate certified robustness like randomized smoothing and bound propagation can be applied to train the secure neural networks (NNs). In the inference stage, pre-processing methods can help mitigate the adversarial perturbations, and adversarial detection can reject the malicious input. The certified exact methods and Lipschitz methods can be utilized as the analysis tools to provide a robustness bound.
\subsection{Empirical Defense}
\indent
\par Empirical defenses improve the robustness of the models against specific adversarial sample attacks through specific methods. Such defenses include adversarial training~\cite{madry2018towards}, modification of model structure~\cite{yan2023wavelet}, de-noising~\cite{xie2019feature}, randomization~\cite{xie2018mitigating}, and other methods to guarantee model robustness. The key advantage of empirical defenses is that they are typically easy to implement and can provide effective defense mechanisms against various attacks. However, the main disadvantages of these approaches are that they usually rely on specific types of attacks and, therefore, may need to be robust enough against unknown attacks or slightly changing forms of attacks. Table \ref{tab:empirical_defense_methods} lists the classical empirical defense methods in the past decade. 
\subsubsection{Adversarial Training}
\indent
\par By far, adversarial training is the most effective way to conduct empirical defense. Such a mechanism can realize data augmentation of adversarial examples in the training process. Moreover, the min-max optimization builds the maximum perturbations and the minimum empirical risks~\cite{madry2018towards} to realize the ``gradient penalty" paradigm, which is beneficial to defend against the gradient-based attacks.
\par Madry et al.~\cite{madry2018towards} propose the first milestone work in adversarial training. However, two open issues are obvious. First, there exists degradation for clean accuracy, while the robust accuracy is not considerable. Another issue is training efficiency that it takes almost two days to train a robust adversarially-trained WideResnet on the CIFAR dataset~\cite{krizhevsky2009learning}. Therefore, most sequential studies of vanilla adversarial training~\cite{madry2018towards} focus on either boosting robust generalization or improving training efficiency.
\par Robust generalization improvement is still a valuable research problem in adversarial training. The Vanilla Adversarial Training method~\cite{madry2018towards} utilizes the white-box PGD method as the default attack strategy. It does not pay attention to the defenses against transfer-based black-box attacks. It may converge to a degenerate global minimum, where small curvature artifacts corrupt the data point and obfuscate the decision of the neural networks. To address this issue, Tram{\`e}r et al.~\cite{tramer2018ensemble} conduct an Ensemble Adversarial Training method that augments the training data with perturbations transferred from other models. This method performs considerably on both CIFAR-10 dataset~\cite{krizhevsky2009learning} and ImageNet dataset~\cite{deng2009imagenet}. Adversarial Diversity Promoting (ADP)~\cite{pang2019improving} and DVERGE~\cite{yang2020dverge} are preeminent successors of this empirical defense method. Adversarial examples can also be regarded as the distribution with the divergence using the distribution of natural examples as a reference. Song et al.~\cite{song2019improving} proposes a paradigm of adversarial training with domain adaptation (ADPA) to boost the robust generalization of neural networks. Since the adversarial perturbations can be regarded as the abnormal noises~\cite{fawzi2016robustness}, the classical non-local denoising method~\cite{buades2005non} can be incorporated in adversarial training to formulate the Feature-Denoising Adversarial Training (FD-AT) framework~\cite{xie2019feature}. Another similar method is ME-Net (Matrix Estimation Net) in which the adversarially-trained neural network leverages matrix estimation (ME) to reconstruct images and mitigate perturbations~\cite{yang2019me}. 
\par The milestone work to improve robust generalization is TRADES~\cite{zhang2019theoretically}. Zhang et al.~\cite{zhang2019theoretically} seek a tradeoff between robustness and accuracy with the proof of a tight differentiable upper bound using the theory of classification-calibrated empirical risks. The TRDES method divides the error against robustness into two components, the estimation error against natural samples and the boundary error, and estimates upper bounds for each of them. In this case, the upper bound estimation for the natural error uses a convex loss function. The upper bound estimation for the boundary error uses a geometric metric based on the loss function, such as KL Divergence.This approach captures the tradeoff between robustness and accuracy of the model well and provides theoretical guarantees. Yan et al.~\cite{yan2024enhance} give proof that the geodesic is the shortest trajectory between two points and propose the Geodesic Adversarial Training (GeodesicAT) framework to enhance the TRADES method. Besides the min-max optimization mechanism, data quantity and quality would also decide the robustness of neural networks. Alayrac et al.~\cite{alayrac2019labels} postulate that the unlabeled data significantly improves robustness and propose an Unsupervised Adversarial Training (UAT) method to deploy a robust machine learning model. The Misclassification Aware Adversarial Training (MART) method~\cite{wang2020improving} is another milestone work that the correctly-classified/incorrectly-classified training samples are regularized in different ways during the adversarial training process. Balunovic et al.~\cite{balunovic2020adversarial} propose a Convex Layerwise Adversarial Training (COLT) method to bridge the gap between adversarial training and provable defense. Chan et al.~\cite{chan2020jacobian} propose the Jacobian Adversarial Regularized Network (JARN) method with the utilization of optimizing the saliency of a classifier's Jacobian by adversarially regularizing the model's Jacobian to resemble natural training images. Then, the method is extended to the frequency domain~\cite{chan2022does} to boost adversarial robustness. Adversarial robustness improvement can also connect with the uncertainty calibration to formulate Confidence-calibrated Adversarial Training (CCAT)~\cite{stutz2020confidence} or return to the tradition of cybernetics to build a robust Close- Loop Control Neural Network (CLC-NN)~\cite{chen2021towards}. The original adversarial training method~\cite{madry2018towards} suffers from the phenomenon of ``robust overfitting"~\cite{tsipras2019robustness}. Many sequential studies have been proposed to alleviate such an issue. Zhang et al.~\cite{zhang2020attacks} prove that the fixed large attack step size may lead the neural network to be immersed in the local optima of robustness and propose a Friendly Adversarial Training (FAT) method based on the curriculum learning mechanism. Furthermore, the Geometry-aware Instance-reweighted Adversarial Training (GAIRAT) method~\cite{zhang2021geometry} adaptively assigns the larger weights to the difficult adversarial examples. Improving the sample efficiency of adversarial examples is another feasible direction in which man can generate the adversarial distributions rather than the point-wise adversarial examples~\cite{dong2020adversarial}. The regularization method is also important in adversarial training, e.g., weight decay and early stopping mentioned in the study of ``Bag of Tricks for AT"~\cite{pang2021bag}, adversarial weight perturbation (AWP)~\cite{wu2020adversarial}, data augmentation~\cite{rebuffi2021data}. Recently, Pang et al.~\cite{pang2022robustness} advocate for the employment of local equivariance as a means to delineate the ideal behavior of a robust model. This approach leads to the formulation of a self-consistent robust error, which they have named SCORE. The new incremental work of adversarial training is boosting the network representation through the wavelet regularization~\cite{yan2023wavelet}, Diffusion-Model-Improves-Adversarial-Training (DM-Improves-AT)~\cite{wang2023better}, ``Learnable Attack Strategy" Adversarial Training (LAS-AT)~\cite{jia2022adversarial} based on the REINFORCE algorithm~\cite{williams1992simple}. In summary, robust generalization improvement is still a core problem of adversarial training. The gap between robust and natural accuracy in a large-scale dataset like ImageNet~\cite{deng2009imagenet} is still huge and deserves further study.
\par Besides robust generalization, training efficiency is another issue in adversarial training research. Shafahi et al.~\cite{shafahi2019adversarial} first attempt to address this problem to eliminate the overhead of adversarial perturbation generation. The Free Adversarial Training method recycles the gradient information during optimization, which can realize adversarial robustness on the ImageNet dataset in a single workstation with 4 P100 Graph Processing Units (GPUs). The training time is only two days. Zhang et al.~\cite{zhang2019you} prove that adversarial training can be regarded as a discrete-time differential game. Based on Pontryagin Maximum Principle (PMP), they have proposed a You Only Propagate Once (YOPO) method, in which the forward and backpropagation can only be restricted within the first layer of the neural network during the model parameter update process. Wong et al.~\cite{wong2020fast} show that it is possible to train empirically robust models with the FGSM method. It further reduces the training time. Qin et al.~\cite{qin2019adversarial} show that promoting linearity can alleviate the gradient obfuscation problem of adversarial training and accelerate the training speed. Improving fast adversarial training is a fascinating research direction. Andriushchenko et al.~\cite{andriushchenko2020understanding} propose a new GradAlign regularization method to alleviate the ``catastrophic overfitting" issue by maximizing the gradient alignment during the attack process. Grabinski et al.~\cite{grabinski2022frequencylowcut} postulate that poor down-sampling operations cause aliasing artifacts and contribute to the adversarial vulnerability of neural networks. Therefore, the proposed FrequencyLowCut pooling method can be combined with the fast FGSM adversarial training method to improve the training efficiency of adversarial defense. The new robust critical fine-tuning method~\cite{zhu2023improving} can enhance robust generalization in non-robust critical modules with light training costs. Overall, the reduction of adversarial training costs is still an open problem.
\par Currently, on large vision datasets like ImageNet~\cite{deng2009imagenet} and Cityscapes~\cite{cordts2016cityscapes}, the gap between robustness and clean accuracy is still huge. It highlights the need for continual research on adversarial training. In the cyber-physical system of automated driving, a fast adversarial training method can ensure both the security and safety of vehicles, which deserves further study.
\subsubsection{Other Training-stage Defense Methods against Adversarial Examples}
\indent
\par Besides adversarial training, other training-stage defense methods are still feasible to improve adversarial robustness.
\par Yan et al.~\cite{yan2018deep} propose a Deep Defense method with the introduction of an adversarial perturbation-based regularization item in the loss function. The ADP method~\cite{pang2019improving} would encourage the diversity of the decision output based on the ensemble mechanism. The stable neural ordinary differential equation (ODE) model~\cite{kang2021stable} is also important to defend against adversarial attacks. The Defense Distillation method~\cite{papernot2016distillation} can provide a shield under gradient-based attacks, while it is vulnerable under adaptive attacks.
\par Overall, the adversarial training methods are the preferred choices for adversarial defenses. However, other techniques like the distillation method~\cite{papernot2016distillation} and neural ODE~\cite{kang2021stable} are still encouraging. The distillation methods have been utilized in the perception model development of automated driving~\cite{sautier2022image,zhou2023unidistill}. Although the distillation method is weak under adaptive attacks, it can be a feasible method against UAPs and other black-box noises.
\subsubsection{Inference-time Defense}
\indent
\par Sometimes, the adversaries launch the attack off-guard. The deployment of an adversarially-trained model would be too late to provide the shield. The inference-time defense can play as an expedient plug-in in the AI software infrastructure. The input transformation is an intuitive defense method~\cite{guo2018countering}, including bit-depth reduction, JPEG compression, total variance minimization, and image quilting. Utilizing the Randomization method~\cite{xie2018mitigating} can also provide the elastic defense which confuses the adversaries. Raff et al.~\cite{raff2019barrage} explore a similar idea of a stochastic Barrage of Random Transformations (BaRT) to defend against adaptive attacks. . One assumption exists that adversarial examples are mainly present in the low probability region of the training distribution. Therefore, Song et al.~\cite{song2018pixeldefend} propose a new method based on generative models, PixelDefend, using statistical hypothesis testing and pixel purification to defend against attacks, building a robust neural network model. Pang et al.~\cite{pang2020mixup} propose a method known as Mixup Inference, which involves blending the input with other random, clean samples. This technique is designed to shrink and transfer the equivalent perturbation if the input is adversarial. Self-supervised Online Adversarial Purification (SOAP)~\cite{shi2021online} is also a novel defense strategy that utilizes the label-independent nature of self-supervised signal to mitigate adversarial perturbations. Building upon the observation that adversaries are generated through iterative minimization of a network's prediction confidence, Alfarra et al.~\cite{alfarra2022combating} design an anti-adversary method to prevent the construction of adversarial examples. Furthermore, since the diffusion model can be utilized for denoising, Nie et al.~\cite{nie2022diffusion} propose an adversarial purification method based on the diffusion model~\cite{song2021score}.
\par The black-box attacks would be more common in the real applications. Qin et al.~\cite{qin2021random} propose a lightweight defense method of random noise defense (RND) for the score-based black box attacks, which adds random noise to each query to interfere with attackers' gradient estimation or random search, thus reducing the attack efficiency. Chen et al.~\cite{chen2022adversarial} propose an Adversarial Attack on Attackers (AAA) method to fool the greedy attackers into incorrect directions by slight perturbations on the neural network outputs in the test time. This method has three advantages: (1) the mitigation of the score-based black-box attacks, (2) the preservation of clean accuracy, and (3) uncertainty calibration. Wang et al.~\cite{wang2021fighting} leverage the defensive entropy minimization (dent) mechanism to output the robust prediction under the white-box, black-box, adaptive attacks on CIFAR-10/100 and ImageNet dataset. The recent Boundary Defense method~\cite{aithal2022boundary} can also guard the MLaaS system, that the model will detect the boundary samples as those with low classification confidence and add white Gaussian noise to their logits. 
\par Compared to the training-time defense methods, the inference-time defense needs less computation costs. It is easy to deploy in real applications such as automated driving and unmanned aerial vehicles. 
\subsubsection{Adversarial Detection}
\indent
\par Adversarial examples can be viewed as anomalous data. The straightforward defense method is to detect them. One questionable view is that the adversarial perturbations are usually not perceptible, and some attacks based on $\ell_{0}$-norm and $\ell_{2}$-norm will limit the changes to pixels. Nevertheless, this type of defense method cannot be neglected.
\par In the early years of adversarial examples research, adversarial detection is a popular method to defend adversarial examples. The binary classifier~\cite{metzen2016detecting}, the SafetyNet based on the Support Vector Machines (SVM)~\cite{lu2017safetynet}, MagNet with diverse separate detector networks and a reformer network based on the manifold assumption~\cite{meng2017magnet} show their considerable performance on adversarial detection. Another detection tools include Gausian Mixture Model (GMM)~\cite{zheng2018robust}, Mahalanobis distance~\cite{lee2018simple}, reverse cross entropy~\cite{pang2018towards}, and local intrinsic property~\cite{ma2018characterizing}. The generative model can also be leveraged to detect adversarial examples. For example, Yang et al.~\cite{yang2021class} build a class-disentanglement variation autoencoder (CD-VAE) to detect adversarial examples.
\par The detection of black-box adversarial examples has received huge attention in recent years. PRADA~\cite{juuti2019prada} is the first detection model to defend the transfer-based black-box attacks. The stateful detection (SD) method~\cite{chen2020stateful} assumes that the attack query sequence exhibits high similarity due to the iterative attack search. The MLaaS system can reject the query and ban the malicious account. The Blacklight method~\cite{li2022blacklight} inherits such an assumption and replaces the $\ell_{2}$ distance metric utilized in the SD work~\cite{chen2020stateful} with the fingerprints.
\par The detection methods have demonstrated superior performance in the black-box defense~\cite{chen2020stateful,li2022blacklight}. Whether deploying it into automated driving is worthwhile is still under scrutiny.
\subsection{Certified Defense}
\begin{table}[!t]
\scriptsize
\centering
\caption{Classical certified defense methods. The main text gives the full names of abbreviations.}
\label{tab:certified_defense_methods}
\begin{tabularx}{\textwidth}{@{} l @{\hspace{5pt}} *{5}{X} @{}}
\toprule
Method & Degree & Category & Large datasets?\\
\midrule
Reluplex~\cite{katz2017reluplex} & Exact & Formal verification & No \\
Huang et al.~\cite{huang2017safety} & Exact & Formal verification & Yes \\
Ehlers et al.~\cite{ehlers2017formal} & Exact & Formal verification & No \\
Cheng et al.~\cite{cheng2017maximum} & Exact & Mixed integer programming & No \\
Cross-Lipschitz regularization~\cite{hein2017formal} & Approximate & Lipschitz continuity & No \\
Xiang et al.~\cite{xiang2018output} & Exact & Mixed integer programming & No \\
Branch-and-Bound~\cite{bunel2018unified} & Exact & Mixed integer programming & No \\
Convex Outer Adversarial Polytope~\cite{wong2018provable} & Approximate & Convex relaxation & No \\
Random Projection~\cite{wong2018provable} & Approximate & Convex relaxation & No \\
Dvijotham et al.~\cite{dvijotham2018dual} & Approximate & Convex relaxation & No \\
Semi-definite Programming~\cite{raghunathan2018semidefinite} & Approximate & Convex relaxation & No \\
ReLUVal~\cite{wang2018formal,wang2018efficient} & Approximate & Bound propagation
& No \\
Fast-Lip~\cite{weng2018towards} & Approximate & Lipschitz continuity
& No \\
LMT~\cite{tsuzuku2018lipschitz} & Approximate & Lipschitz continuity
& No \\
Richards et al.~\cite{richards2018lyapunov} & Approximate & Cybernetics
& No \\
Fazlyab et al.~\cite{fazlyab2019efficient} & Approximate & Lipschitz continuity
& No \\
GeoCert~\cite{jordan2019provable} & Approximate & Convex relaxation & No \\
Salman et al.~\cite{salman2019convex} & Approximate & Convex relaxation & No \\
IBP~\cite{gowal2019scalable}  & Approximate & Bound propagation & Yes \\
Lipschitz norm ball~\cite{croce2019provable,croce2020provable} & Approximate & Lipschitz continuity & No \\
PixelDP~\cite{lecuyer2019certified} & Approximate &  Randomized smoothing & Yes \\
Vanilla Randomized Smoothing~\cite{cohen2019certified} & Approximate &  Randomized smoothing & Yes \\
Salman et al.~\cite{salman2019provably} & Approximate &  Randomized smoothing & Yes \\
Mangal et al.~\cite{mangal2019robustness} & Approximate &  Uncertainty & No \\
PROVEN~\cite{weng2019proven} & Approximate &  Uncertainty & No \\
Fazlyab et al.~\cite{fazlyab2019probabilistic} & Approximate &  Uncertainty & No \\
Wang et al.~\cite{wang2019verification} & Approximate & Cybernetics & No \\
Wang et al.~\cite{wang2019resnets} & Approximate & Cybernetics & No \\
Chiang et al.~\cite{chiang2020certified}  & Approximate & Bound propagation & No \\
Sparse polynomial optimization~\cite{latorre2020lipschitz} & Approximate & Lipschitz continuity & No \\
Jordan et al.~\cite{jordan2020exactly} & Approximate & Lipschitz continuity & No \\
F-Divergence Smooth~\cite{dvijotham2020framework} & Approximate & Randomized smoothing & Yes \\
$\ell_{\infty}$-distance~\cite{zhang2021towards,zhang2022boosting} & Approximate & Lipschitz continuity & No\\
PointGuard~\cite{liu2021pointguard} & Approximate & andomized smoothing & No\\
GCP-CROWN~\cite{zhang2022general}  & Approximate & Bound propagation & No \\
SortNet~\cite{zhang2022rethinking} &  Approximate & Lipschitz continuity & Yes \\
Schuchardt et al.~\cite{schuchardt2022invariance} & Approximate & Randomized smoothing & Yes \\
Li et al.~\cite{li2022double} & Approximate & Randomized smoothing & Yes \\
LipsFormer~\cite{qi2023lipsformer} &  Approximate & Lipschitz continuity & Yes \\
3deformrs~\cite{schuchardt2022invariance} & Approximate & Randomized smoothing & Yes \\
Alfarra et al.~\cite{alfarra2022data} & Approximate & Randomized smoothing & Yes \\
Anderson et al.~\cite{anderson2022towards,anderson2022certified} & Approximate & Randomized smoothing & Yes \\
Pfrommer et al.~\cite{pfrommer2023projected} & Approximate & Randomized smoothing & Yes \\
\bottomrule
\end{tabularx}
\end{table}
\indent
\par The empirical defense method would meet the challenges of sophisticated adaptive attackers~\cite{athalye2018obfuscated,tramer2020adaptive}. The continuous arms between attackers and defenders motivate a theoretical interpretation of adversarial robustness. The certified defense methods respond to these commands. It consists of a robustness verification approach providing the lower bound of robustness under any attacker without the specification of perturbation type and corresponding robust training methods. This subsection reviews the certified defense methods. Table \ref{tab:certified_defense_methods} lists the classical certified defense methods in the field of robustness research.
\subsubsection{Formal Verification Methods}
\indent
\par The formal verification methods are used to formally verify the robustness of a model to specific input perturbations through solvers or theorem-proving techniques. It is an exact but computationally expensive method. Katz et al.~\cite{katz2017reluplex} focus on the non-convex Rectified Linear Unit (ReLU) activation function, an important ingredient in CNNs. The scalable simplex method of linear programming named Reluplex supports the ReLU constraint. The cost of exactness is a
high computation budget, which limits the usage of Reluplex in real applications. Huang et al.~\cite{huang2017safety} utilize the Satisfiability Modulo Theory (SMT) to provide a verification framework on several datasets, including MNIST~\cite{lecun1998mnist}, CIFAR10~\cite{krizhevsky2009learning}, GTSRB~\cite{stallkamp2011german}, and ImageNet~\cite{deng2009imagenet}. Ehlers et al.~\cite{ehlers2017formal} provide the global linear bound for the piece-wise feed-forward neural networks and reduce the computation cost of SMT on the small dataset. Overall, the formal verification method can provide a relatively exact bound. However, these methods suffer from the high computation cost, which is not applicable for large and foundation models~\cite{bommasani2021opportunities}.
\subsubsection{Mixed Integer Programming Methods}
\indent
\par The utilization of mixed integer programming methods also provides a potential solution to the robust verification of neural networks. Cheng et al.~\cite{cheng2017maximum} quantize the maximum perturbation bound via the mixed integer programming (MIP) method and apply the method in the agent game for safety-critical applications like automated driving. The reachability analysis can be built for both feed-forward ReLU neural networks~\cite{lomuscio2017approach} and Multi-Layer Perception (MLP)~\cite{xiang2018output}. The Branch-and-Bound method can also be used to build a unified view of verification on the small neural networks~\cite{bunel2018unified}. None of these methods have been applied to the large-scale scenarios.
\subsubsection{Convex Relaxation Methods}
\indent
\par Although most neural networks handle non-convex optimization problems, the convex relaxation methods can be utilized in the certified defense framework. Compared to exact methods, convex relaxation methods are approximate numerical optimization techniques for solving convex problems. Convex relaxation methods find the global optimal solution by gradually relaxing the original constraints of the problem into a series of approximate convex optimization sub-problems. Wong et al.~\cite{wong2018provable} utilize a convex outer approximation of the reachable sets. It is shown that the dual problem of convex outer adversarial polytope can represent the backpropagation of DNNs. The Random Projection method~\cite{wong2018provable} scales the provable defense method to the complex scenario. Inspired by the idea of duality, Dvijotham et al.~\cite{dvijotham2018dual} formalize the problem of certified defense as an unconstrained convex optimization problem and obtain a provable robust boundary by solving a Lagrangian relaxation of this unconstrained convex optimization problem. Sub-gradient methods can solve this computing process of the robust boundary. Raghunathan et al.~\cite{raghunathan2018semidefinite} propose a Semi-definite Relaxation method to solve the max-cut problem for a certified robustness bound. The sequential improvement work of convex relaxation methods can be built on the polyhedral complices~\cite{jordan2019provable}. Furthermore, Salman et al.~\cite{salman2019convex} propose a tight convex relaxation barrier method in a hierarchical framework. Briefly, the convex relaxation methods are approximation methods with high efficiency. However, the scale of these methods on the large-scale datasets is difficult.
\subsubsection{Bound Propagation Methods}
\indent
\par The bound propagation method is a conservative approximation method to simplify the robust verification process with the computation by calculating the interval range of the output over the input at each layer. Gowal et al.~\cite{gowal2019scalable} propose a method called Interval Bound Propagation (IBP) to compute the worst-case bounds of the network output by propagating upper and lower bounds on the activation values at each layer. By optimizing the network parameters, the bounds of the network output are made to satisfy the given specification. The ReLUVal method~\cite{wang2018formal,wang2018efficient} using symbolic intervals can provide a tight formal security and safety bound for neural networks. Chiang et al.~\cite{chiang2020certified} abstract a certified defense model and resolve the problem with the IBP method. Zhang et al.~\cite{zhang2022general} generalize the bound propagation method with the general cutting plane (GPC) to realize robust verification in the GCP-CROWN framework. 
\subsubsection{Lipschitz Continuity Methods}
\indent
\par Lipschitz continuity is a concept of stability analysis. Specifically, a function is said to be Lipschitz continuous on a domain if there exists a real constant $L \geq 0$, known as a Lipschitz constant, such that for the data $x_{1}$, $x_{2}$, the function output change satisfies the inequality:
\begin{equation}\label{eq:lipschitz}
\left|f\left(x_{1}\right)-f\left(x_{2}\right)\right| \leq L\left|x_{1}-x_{2}\right|,
\end{equation}
for all $x_{1}$ and $x_{2}$ in the domain $f$.
\par Hein et al.~\cite{hein2017formal} firstly scale the Lipschitz continuity concept to the certified robustness problem and propose a Cross-Lipschitz regularization method for the defense. The Lipschitz continuity tools can be built on the spheres with norm balls to provide a formal robustness guarantee that does not depend on the space size~\cite{croce2019provable,croce2020provable}. Weng et al.~\cite{weng2018towards} propose a fast Lipschitz (Fast-Lip) method on the MNIST~\cite{lecun1998mnist} and CIFAR~\cite{krizhevsky2009learning} datasets with the speed acceleration. Tsuzuku et al.~\cite{tsuzuku2018lipschitz} propose a training method called Lipschitz-Margin Training (LMT), which improves the certified robustness of the neural networks by calculating an upper bound on the Lipschitz constant of each component and using that upper bound to train the network with looser robust bounds. Fazylab et al.~\cite{fazlyab2019efficient} show the activation functions can be interpreted as gradients of convex potential functions and calculate the Lipschitz constant with the semi-definite programming method. Jordan et al.~\cite{jordan2020exactly} calculate the non-smooth vector-valued functions via the norm of the generalized Jacobian. Latorre et al.~\cite{latorre2020lipschitz} implement the Lipschitz constant estimation for certified robustness via the sparse polynomial optimization mechanism. The Lipschitz constant can also be calculated via the $\ell_{\infty}$-distance~\cite{zhang2021towards,zhang2022boosting} or the boolean function~\cite{zhang2022rethinking}. Qi et al.~\cite{qi2023lipsformer} scale the Lipschitz continuity method to the vision transformer model (LipsFormer) and replace the unstable LayerNorm model with the Lipschitz continuous CenterNorm module. 
\par Generally speaking, the Lipschitz continuity methods can handle the neural networks with a non-differentiable input transformation, which is suitable for analyzing the activation function in deep learning. However, the Lipschitz continuity methods would be inclined to output the looser robust bound. 
\subsubsection{Randomized Smoothing Methods}
\indent
\par The randomized smoothing methods improve the robustness of the model to perturbations by adding random noises around the input data and averaging multiple noisy perturbations of the model. These approaches provide probabilistic robustness guarantees.
\par Lecuyer et al.~\cite{lecuyer2019certified} propose the Pixel Differential Privacy (PixelDP) method, a novel provable defense mechanism against adversarial examples attacks in a specific range. The method is based on the concept of differential privacy, which can be applied to any type of deep neural network and can be applied to large-scale networks and datasets. The main idea of PixelDP is to add noise to the training and prediction time to increase robustness while maintaining provable privacy protection. The Neyman-Pearson lemma~\cite{scott2005neyman} provides a relatively accurate boundary for binary hypothesis testing (robust or vulnerable). Cohen et al.~\cite{cohen2019certified} propose a robust certified defense method under the $\ell_{\infty}$-norm attacks based on the Neyman-Pearson lemma. Besides, Cohen et al.~\cite{cohen2019certified} also show that the Monte Carlo algorithm can evaluate the prediction trustworthiness of smooth classifiers. Salman et al.~\cite{salman2019provably} design an adaptive attack mechanism for the randomized smooth classifier, which provides a robustness guarantee under strong attacks. Dvijotham et al.~\cite{dvijotham2020framework} prove the robustness of smoothed classifiers via the tools of F-Divergence. Randomized smoothing would cause several hidden costs that shrink the decision boundaries with the adoption of the prediction rules~\cite{mohapatra2021hidden}. Moreover, the augmented perturbations do not necessarily solve the boundary shrinkage problem. However, it can help the application of these methods on large-scale datasets. The surprising utilization is on the point-cloud datasets including PointGuard~\cite{liu2021pointguard}, invariance-aware randomized smoothing certificate~\cite{schuchardt2022invariance}, 3Deform Randomized Smoothing (3deformrs)~\cite{perez20223deformrs}. Other incremental studies of randomized smoothing include double sampling randomized smoothing~\cite{li2022double}, projected randomized smoothing~\cite{pfrommer2023projected}, data-dependent randomized smoothing~\cite{alfarra2022data}, optimal randomized smoothing via the semi-infinite linear programming~\cite{anderson2022towards}, locally-biased randomized smoothing~\cite{anderson2022certified}.
\par In summary, the randomized smoothing methods are the acknowledged feasible methods that can be adapted to large-scale datasets. Nevertheless, the shrinking boundary accompanied by smoothing is an open issue. There is still ample room for further research and development in the field of random smoothing.
\subsubsection{Uncertainty-based Methods}
\indent
\par Certified robustness can connect with uncertainty quantification. Mangal et al.~\cite{mangal2019robustness} introduce a novel concept of robustness termed probabilistic robustness, necessitating that the neural network exhibits robustness with a probability of at least 1-$\varepsilon$ concerning the input distribution. This probabilistic approach is pragmatic and offers a systematic method for assessing the robustness of a neural network. The PROVEN method~\cite{weng2019proven} realizes probabilistic robustness in the case of adversarial perturbations that follow a specific probability distribution, providing probabilistic guarantees that the top-1 predictions of the model will not change in a statistically significant sense of verifiable robustness. Fazlyab et al.~\cite{fazlyab2019probabilistic} compute a confidence ellipsoid for the output via semi-definite programming. The uncertainty-based method is intuitive, and can provide a loose robustness bound. However, its assumption of the noise distribution is not always satisfactory in real applications.
\subsubsection{Cybernetics-based Methods}
\indent
\par Modern AI methods can also return to the cybernetics tradition. Therefore, the certified defense methods can also be combined with the robust control framework~\cite{wang2019verification}, the optimal transport algorithm built on the Feynman-Kac Formalism~\cite{wang2019resnets}, and the Lyapunov function~\cite{richards2018lyapunov}. However, the effectiveness of these methods needs further inspection.
\section{Adversarial Examples in Perception Systems of Automated Driving}
\indent
\par The existence of adversarial examples poses a security threat to autonomous driving. This section reviews the significant progress of adversarial examples in automated driving. 
\subsection{Objection Detection}
\indent
\par Object detection is an important task in automated driving. The object detectors based on CNNs~\cite{redmon2016you} and transformers~\cite{carion2020end} can handle most scenarios in automated driving. However, due to the uncertainty of prediction and potential security issues, it is still far away from trustworthiness. For example, many studies show that object detectors are vulnerable to adversarial examples.
\par Xie et al.~\cite{xie2017adversarial} show that both segmentation and detection models will classify multiple objects in an image. Therefore, the attacks can aim at pixels and proposals. Based on the assumption, the Dense Adversary Generation (DAG) method~\cite{xie2017adversarial} attacks the object detection and semantic segmentation models. Wei et al.~\cite{wei2019transferable} propose a method to manipulate the feature maps extracted and improve the transferability on the adversarial examples of object detection. Adversaries can manipulate the patch to defraud the object detectors~\cite{wu2020dpattack,zhao2020object,huang2021rpattack}. Huang et al.~\cite{huang2023t} propose a novel single-model-based black-box adversarial attack method to improve the transferability of attacks against the object detection models. The method is mainly based on a self-ensemble strategy, which includes integrating input data, an attacked model, and an adversarial patch to enhance the transferability of the adversarial patch. Several methods have been proposed to defend against the attacks, including the multi-task learning~\cite{zhang2019towards}, class-aware robust optimization~\cite{chen2021class}, and adversarially-aware convolution module to disentangle gradients for optimization on clean and adversarial
data~\cite{dong2022adversarially}. Achieving adversarial robustness for object detection in automated driving is still a long way to go. 
\subsection{Semantic Segmentation}
\indent
\par In automated driving, semantic segmentation is a critical task that involves correlating each pixel in an image to a specific category label. This process aims to divide the image into areas with a clear semantic meaning to help autonomous driving systems understand and parse the road environment. Through semantic segmentation, automated vehicles can accurately identify various elements of roads, pedestrians, vehicles, and traffic signs and assign them the corresponding labels. In this way, vehicles can make decisions based on the label information, such as avoiding pedestrians and vehicles, obeying traffic rules, etc. Currently, the recognition performance of the semantic segmentation model is satisfactory~\cite{chen2018encoder,xie2021segformer,jain2023oneformer}. However, the adversarial attacks will cause the degradation of the Mean Intersection over Union (mIoU).
\par After the proposal of DAG~\cite{xie2017adversarial}, the vulnerability problem of dense pixel classification in segmentation still needs to be addressed. Gu et al.~\cite{gu2022segpgd} propose a segmentation-specific PGD called SegPGD. The adversarial training mechanism based on SegPGD will boost the adversarial robustness of the semantic segmentation model. Yin et al.~\cite{yin2022adversarial} give a systematic evaluation of adversarial robustness for CNN-based semantic segmentation models in automated driving. The ViT-based semantic segmentation models have become the mainstream~\cite{xie2021segformer,jain2023oneformer} in automated driving. The robustness study of these models is still an open problem.
\subsection{Adversarial Examples in 3D Perception}
\indent
\par LiDAR (Light Detection and Ranging) technology plays a pivotal role in the advancement of automated driving systems, offering superior capabilities in obstacle detection, localization, and navigation compared to traditional camera sensors. LiDAR devices boast higher resolution ratios for distance, angle, and speed, coupled with robust anti-interference properties, making them particularly effective under adverse weather conditions. Many contemporary commercial autonomous vehicles leverage systems that integrate LiDAR with camera devices for enhanced perception~\cite{li2023key}. Architectures relying solely on LiDAR~\cite{lang2019pointpillars,shi2020pv,shi2023pv} as well as those employing sensor fusion techniques~\cite{li2022deepfusion,bai2022transfusion} have demonstrated significant achievements on academic benchmarks such as KITTI~\cite{geiger2012we} and nuScenes~\cite{caesar2020nuscenes}.
\par Despite these advancements, security research has illuminated vulnerabilities within LiDAR-based systems, revealing potential for spoofed attacks and the generation of malicious obstacles~\cite{cao2019adversarial,sun2020towards,tu2020physically,yang2021robust}. These adversarial challenges extend to multi-sensor fusion models as well, exposing similar susceptibilities~\cite{cao2021invisible,wang2022adversarial}. In response, several exhaustive investigations have endeavored to assess the robustness of LiDAR-based 3D object detection~\cite{kong2023robo3d,zhang2023comprehensive} and sensor fusion models~\cite{dong2023benchmarking}. However, as the landscape of foundational and planning-oriented models evolves~\cite{liu2023segment,hu2023planning}, the focused exploration into their adversarial robustness becomes increasingly significant.
\subsection{Trajectory Prediction}
\indent
\par Trajectory prediction stands as a pivotal component in the applications of automated driving, tasked with forecasting the movements of nearby vehicles and pedestrians to prompt the control, planning, and navigation strategies. While these models have demonstrated impressive efficacy in naturalistic settings~\cite{salzmann2020trajectron++,yuan2021agentformer,cui2019multimodal}, their susceptibility to adversarial attacks poses a severe challenge~\cite{cao2022advdo,zhang2022adversarial,tan2023targeted}. In response, diverse defensive strategies have been advanced, such as domain-specific data augmentation and adversarial training~\cite{cao2023robust}, semi-supervised semantics-guided adversarial training, and adversarial defenses that leverage causal Total Direct Effect (TDE) inference~\cite{duan2023causal}. Currently, many pieces of research still focus on the white-box attack scenarios. However, most automated driving systems are black-box MLaaS systems. Therefore, the research community needs to allocate more attention to the black-box adversarial robustness of trajectory prediction models.
\section{Adversarial Examples and SOTIF}
\indent \par Driving safety, traffic efficiency, and low-carbon transportation are several significant factors for automated vehicles. The driving safety includes functional safety, cybersecurity, and SOTIF. Usually, the research of adversarial examples belongs to the category of cybersecurity. Currently, there are three critical challenges in SOTIF~\cite{li2023key}: the long-tailed scenario problem, the system complexity and diversity in automated driving, and the AI algorithm's inexplicability and uncertainty. All these issues connect with the adversarial robustness of automated vehicles. Firstly, the existence of adversarial obstacles and malicious traffic signs plays a critical role in the operational environment of self-driving cars and thus cannot be overlooked. Secondly, while overparameterization has been shown to enhance the adversarial robustness of neural networks~\cite{bubeck2021universal}, there is a pressing need for practical systems to adopt lightweight yet robust models~\cite{ye2019adversarial}. Thirdly, the relationship between the uncertainty inherent in AI methodologies and adversarial robustness has been the subject of several studies~\cite{smith2018understanding,tuna2022closeness,schweighofer2023quantification}, but all these studies have not been extended to automated driving scenarios. The offline safety design, online safety monitoring, and active ongoing learning~\cite{li2023key} should take adversarial robustness into consideration. Moreover, the universal adversarial examples~\cite{hendrycks2019benchmarking,yan2023exploring} offers a unique opportunity to simulate adverse weather conditions and sensor failures in automated driving, serving as practical test cases for AI system evaluation and enhancement through active learning. By reinterpreting external security threats as catalysts for improving internal system safety, we can shift the paradigm towards a more resilient automated driving ecosystem. Despite their significance, these areas of study have yet to garner the attention they merit. Moving forward, they represent critical avenues for in-depth exploration by the research community.
\section{Future Research Directions}
\indent \par There are several research directions related to adversarial robustness in automated driving.
\par Firstly, the landscape of foundational models in computer vision has witnessed significant advancements in recent years. Among these, the Segment-Anything model~\cite{kirillov2023segment} demonstrates the capability for zero-shot recognition across various application domains, though its efficacy and robustness within the context of automated driving remain to be fully explored. Furthermore, recent innovations have introduced a novel architectural framework that facilitates sequential modeling over linear time through the utilization of selective state spaces~\cite{gu2023mamba,zhu2024vision}, effectively addressing the computational challenges associated with processing long sequences by Transformers. This development heralds the emergence of a new avenue for research.
\par Secondly, the online automated driving algorithms need to run on specific chips rather than NVIDIA A100 GPUs. Therefore, addressing the adversarial robustness of the edge computing scenarios is important. Some recently published work can be the reference for the robustness related to the compression and quantization~\cite{ye2019adversarial,gui2019model,xiao2023robustmq}.
\par Thirdly, elucidating the nexus between adversarial robustness and prediction uncertainty emerges as a critical endeavor. Additionally, the PixelDP method~\cite{lecuyer2019certified} offers a pioneering approach that bridges the domains of privacy and robustness. Within the realm of automated driving, the integration of these three pivotal elements, adversarial robustness, prediction uncertainty, and privacy into a cohesive and responsible AI framework, is imperative for comprehensive risk management.
\par Last but not least, SOTIF evaluation and improvement through the tools of adversarial robustness are another vital way to ensure the safety of automated driving systems. SOTIF underscores the criticality of employing AI technologies and verification strategies adeptly to identify, mitigate, and manage emergent risks. This process necessitates a rigorous analysis of the system's intended functionality, the identification of potential hazard scenarios, and the deployment of measures aimed at mitigating these identified threats. The goal is to ensure that the system is designed and implemented in strict adherence to the safety requirements, with residual risks diminished to an acceptable threshold. For instance, the application of Universal Adversarial Perturbations (UAPs) can critically assess the safety of perception systems within autonomous vehicles. Furthermore, methodologies such as rapid adversarial training techniques~\cite{wong2020fast,andriushchenko2020understanding} can serve as the potential tools within automated driving systems to address immediate risks and enhance model SOTIF. Additionally, certified randomized smoothing methods~\cite{cohen2019certified,li2022double,liu2021pointguard} provide a certified robustness $\varepsilon$-region, offering a foundational model for the development of secure and safe systems within the ICV industry.
\section{Conclusion}
\indent \par This survey comprehensively examines the evolution of research on adversarial robustness over the past decade, highlighting the key contributions pertinent to the automated driving systems. Furthermore, it identifies several prospective research achievements, offering preliminary insights aimed at addressing these emergent challenges. Our review synthesizes the existing scholarship with the forward-looking perspectives, positioning itself as a pivotal resource for stakeholders in cybersecurity and SOTIF within the realm of automated driving. Notably, this work does not delve into the theoretical foundations of adversarial vulnerability and robustness in deep learning frameworks, designating this critical area as a subject for future inquiry.
\scriptsize
\bibliographystyle{unsrt}
\bibliography{bookchapter}

\end{document}